%% file: main.tex
\documentclass[10pt,twocolumn,letterpaper]{article}

\usepackage{iccv}              

\input{preamble}

\definecolor{iccvblue}{rgb}{0.21,0.49,0.74}
\usepackage[pagebackref,breaklinks,colorlinks,allcolors=iccvblue]{hyperref}
\usepackage{mathtools}

\def\paperID{00101} 
\def\confName{ICCV}
\def\confYear{2025}

\title{Joint Self-Supervised Video Alignment and Action Segmentation}

\author{Ali Shah Ali$^\dagger$~~~~~Syed Ahmed Mahmood$^\dagger$~~~~~Mubin Saeed~~~~~Andrey Konin\\M. Zeeshan Zia~~~~~Quoc-Huy Tran\\
\\
Retrocausal, Inc., Redmond, WA\\
\url{www.retrocausal.ai}
}

\begin{document}
\maketitle

\begin{abstract}
We introduce a novel approach for simultaneous self-supervised video alignment and action segmentation based on a unified optimal transport framework. In particular, we first tackle self-supervised video alignment by developing a fused Gromov-Wasserstein optimal transport formulation with a structural prior, which trains efficiently on GPUs and needs only a few iterations for solving the optimal transport problem. Our single-task method achieves the state-of-the-art performance on multiple video alignment benchmarks and outperforms VAVA, which relies on a traditional Kantorovich optimal transport formulation with an optimality prior. Furthermore, we extend our approach by proposing a unified optimal transport framework for joint self-supervised video alignment and action segmentation, which requires training and storing a single model and saves both time and memory consumption as compared to two different single-task models. Extensive evaluations on several video alignment and action segmentation datasets demonstrate that our multi-task method achieves comparable video alignment yet superior action segmentation results over previous methods in video alignment and action segmentation respectively. Finally, to the best of our knowledge, this is the first work to unify video alignment and action segmentation into a single model. Our code is available on our research website: \url{https://retrocausal.ai/research/}.
\end{abstract}

\input{Sections/introduction.tex}
\input{Sections/relatedwork.tex}
\input{Sections/method.tex}
\input{Sections/experiments.tex}
\input{Sections/conclusion.tex}

\appendix
\input{supp.tex}

{
\small
\bibliographystyle{ieeenat_fullname}
\bibliography{references}
}

\end{document}

%% file: preamble.tex
\usepackage{graphicx}
\usepackage{booktabs}
\usepackage{amsmath}
\usepackage{amssymb}

\usepackage{times}
\usepackage{epsfig}
\usepackage{comment}
\usepackage{epigraph}
\usepackage{multirow}
\usepackage{hhline}
\usepackage{makecell}
\usepackage{bm}
\usepackage{subcaption}
\usepackage{soul}
\usepackage{array}
\usepackage{color,colortbl}

\makeatletter
\def\blfootnote{\gdef\@thefnmark{}\@footnotetext}
\makeatother

\definecolor{beaublue}{rgb}{0.74, 0.83, 0.9}

\newcommand{\argmin}{\operatornamewithlimits{argmin}}
\newcommand{\argmax}{\operatornamewithlimits{argmax}}

%% file: Sections/introduction.tex
\section{Introduction}
\label{sec:introduction}
{\blfootnote{$^{\dagger}$ indicates joint first author.\\ \{alishah,ahmed,mubin,andrey,zeeshan,huy\}@retrocausal.ai.}}

\begin{figure}[t]
	\centering
		\includegraphics[width=1.0\linewidth, trim = 0mm 25mm 75mm 0mm, clip]{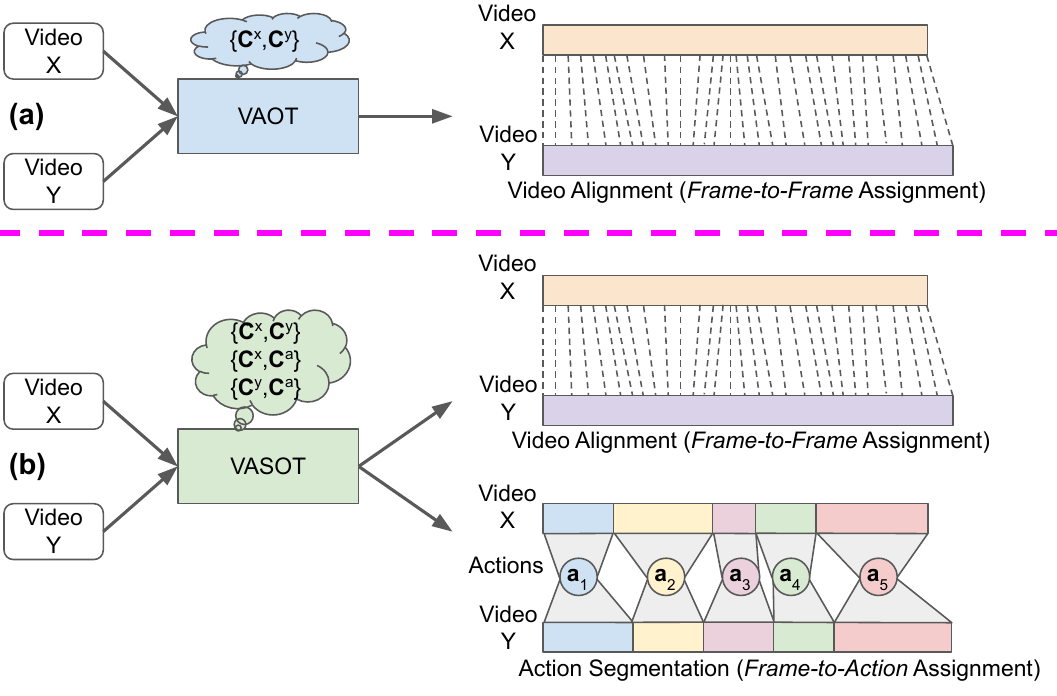}
	\caption{(a) Our self-supervised video alignment method (VAOT) based on a fused Gromov-Wasserstein optimal transport with structural priors $\{\textbf{C}^x,\textbf{C}^y\}$. (b) Our joint self-supervised video alignment and action segmentation method (VASOT) based on a unified optimal transport with structural priors $\{\textbf{C}^x,\textbf{C}^y\}$ for video alignment and $\{\textbf{C}^x,\textbf{C}^a\}$ and $\{\textbf{C}^y,\textbf{C}^a\}$ for action segmentation.}
	\label{fig:teaser}
\end{figure}

Though the past decade has witnessed remarkable progress in human activity understanding in videos, the majority of the research efforts have been invested in action recognition~\cite{carreira2017quo,tran2018closer,wang2018non,feichtenhofer2019slowfast}, which categorizes simple actions in short videos. In this paper, we study the two less-explored problems, i.e., temporal video alignment (\emph{frame-to-frame} assignment), which establishes framewise correspondences between long videos recording a complex activity, and temporal action segmentation (\emph{frame-to-action} assignment), which assigns frames of long videos capturing a multi-phase activity to phase/action labels. Since acquiring per-frame annotations for supervised training is generally difficult and costly, we are interested in self-supervised approaches for video alignment and action segmentation.

One popular group of self-supervised video alignment methods rely on global alignment techniques widely used in time series literature. For example, LAV~\cite{haresh2021learning} utilizes dynamic time warping~\cite{cuturi2017soft} by assuming monotonic orderings and no background/redundant frames. VAVA~\cite{liu2022learning} relaxes the above assumptions by incorporating an optimality prior into a standard Kantorovich optimal transport framework~\cite{cuturi2013sinkhorn}, along with an inter-video contrastive term and an intra-video contrastive term. However, it is challenging to balance multiple losses as well as handle repeated actions. Similarly, self-supervised action segmentation methods based on optimal transport have been introduced, including TOT~\cite{kumar2022unsupervised} and UFSA~\cite{tran2024permutation} which may suffer in cases of order variations, unbalanced segmentation, and repeated actions. ASOT~\cite{xu2024temporally} addresses these drawbacks via a fused Gromov-Wasserstein optimal transport framework with a structural prior, which outperforms previous works in self-supervised action segmentation. Lastly, though both self-supervised video align and action segmentation require fine-grained temporal understanding of videos, their interaction in a multi-task learning setup has not been explored.

Motivated by the above observations, we first propose \emph{VAOT} (see Fig.~\ref{fig:teaser}(a)), a novel self-supervised video alignment approach based on a fused Gromov-Wasserstein optimal transport formulation with a structural prior, which tackles order variations, background/redundant frames, and repeated actions in a single global alignment framework. Our single-task model trains efficiently on GPUs and requires few iterations to derive the optimal transport solution, while outperforming previous methods, including VAVA~\cite{liu2022learning}, on video alignment datasets. Moreover, we develop \emph{VASOT} (see Fig.~\ref{fig:teaser}(b)), a joint self-supervised video alignment and action segmentation approach by exploring the relationship between self-supervised video alignment and action segmentation via a unified optimal transport framework. Our multi-task model performs on par with prior works on video alignment benchmarks yet establishes the new state of the art on action segmentation benchmarks. In addition, our joint model requires training and storing a single model and saves both time and memory consumption as compared to two separate single-task models. Lastly, we observe in Sec.~\ref{sec:sota_comparison} that, in a multi-task learning setting, action segmentation provides little boost to video alignment results, whereas video alignment increases action segmentation performance significantly.

In summary, our contributions include:
\begin{itemize}
    \item We propose a fused Gromov-Wasserstein optimal transport formulation with a structural prior for self-supervised video alignment, outperforming previous methods. Our single-task method learns efficiently on GPUs, needing few iterations to obtain the optimal transport solution.
    \item We develop a unified optimal transport-based approach for simultaneous self-supervised video alignment and action segmentation, yielding comparable video alignment but superior action segmentation results over previous methods. Our joint approach requires training and storing a single model, saving both time and memory usage.
    \item We conduct extensive experiments on several video alignment and action segmentation datasets, i.e., Pouring, Penn Action, IKEA ASM, 50 Salads, YouTube Instructions, Breakfast, and Desktop Assembly, to validate the advantages of our single-task and multi-task methods. To our best knowledge, our work is the first to combine video alignment with action segmentation.
\end{itemize}

%% file: Sections/relatedwork.tex
\section{Related Work}
\label{sec:relatedwork}

\noindent \textbf{Self-Supervised Learning.}
Early self-supervised learning methods focus on designing image-based pretext tasks with pseudo-labels as supervision signals for learning representations such as image colorization~\cite{larsson2016learning,larsson2017colorization}, object counting~\cite{noroozi2017representation,liu2018leveraging}, predicting rotations~\cite{gidaris2018unsupervised}, solving puzzles~\cite{carlucci2019domain,kim2019self}, image inpainting~\cite{jenni2020steering}, and image clustering~\cite{caron2018deep,caron2019unsupervised}. The above image-based approaches mostly extract spatial cues from the image content. Recently, significant efforts have been invested in video-based self-supervised learning methods, which exploit both spatial and temporal information. Examples of video-based pretext tasks include forecasting future frames~\cite{srivastava2015unsupervised,vondrick2016generating,ahsan2018discrimnet,diba2019dynamonet}, enforcing temporal coherence~\cite{mobahi2009deep,zou2011unsupervised,goroshin2015unsupervised}, predicting temporal order~\cite{misra2016shuffle,lee2017unsupervised,fernando2017self,xu2019self}, arrow of time~\cite{pickup2014seeing,wei2018learning}, and pace~\cite{benaim2020speednet,wang2020self,yao2020video}, and utilizing contrastive learning~\cite{feichtenhofer2021large,hu2021contrast,qian2021spatiotemporal,dave2022tclr}. More recently, skeleton-based self-supervised learning methods with skeleton-based pretext tasks, e.g., skeleton inpainting~\cite{zheng2018unsupervised}, motion prediction~\cite{su2020predict}, skeleton sequence alignment~\cite{kwon2022context,tran2024learning}, and utilizing neighborhood consistency~\cite{si2020adversarial}, motion continuity~\cite{su2021self}, and multiple pretext tasks~\cite{lin2020ms2l}, have been introduced. These skeleton-based approaches may suffer from human pose estimation errors and missing context details. Here, we leverage video alignment and/or action segmentation as our video-based pretext tasks.

\noindent \textbf{Video Alignment.}
Self-supervised video alignment has attracted a great amount of research interest in recent years. TCC~\cite{dwibedi2019tcc} enforces cycle consistencies between corresponding frames across videos for learning representations. Recently, GTCC~\cite{donahue2024learning} extends TCC~\cite{dwibedi2019tcc} by proposing multi-cycle consistencies for tackling repeated actions. Both TCC~\cite{dwibedi2019tcc} and GTCC~\cite{donahue2024learning} perform local alignment by aligning each frame separately. Motivated by global alignment techniques for time series, methods which align the video as a whole have been introduced. LAV~\cite{haresh2021learning} which assumes monotonic orderings and no background/redundant frames relies on dynamic time warping~\cite{cuturi2017soft}. To handle non-monotonic orderings and background/redundant frames, VAVA~\cite{liu2022learning} employs a traditional Kantorovich optimal transport formulation~\cite{cuturi2013sinkhorn} with an optimality prior. In this work, we propose a fused Gromov-Wasserstein optimal transport formulation with a structural prior which handles order variations, background/redundant frames, and repeated actions in a single global alignment framework. In addition, we develop a unified optimal transport-based approach for joint video alignment and action segmentation. 

\noindent \textbf{Action Segmentation.}
Initial works in self-supervised action segmentation perform representation learning and offline clustering as disjoint steps. Please see Ding et al.~\cite{ding2023temporal} for a recent survey. CTE~\cite{kukleva2019unsupervised} trains a temporal embedding first and then employs K-Means to cluster the embedded representations. To enhance CTE~\cite{kukleva2019unsupervised}, VTE~\cite{vidalmata2021joint} and ASAL~\cite{li2021action} introduce a visual embedding and an action embedding respectively. The above methods separate representation learning and offline clustering, prohibiting effective communications between the two modules. Recently, methods which jointly conduct representation learning and online clustering have been developed, e.g., UDE~\cite{swetha2021unsupervised}, TOT~\cite{kumar2022unsupervised}, and UFSA~\cite{tran2024permutation}. However, their performance may deteriorate in cases of order variations, unbalanced segmentation, and repeated actions. To overcome these limitations, ASOT~\cite{xu2024temporally} introduces a fused Gromov-Wasserstein optimal transport formulation with a structural prior. Here, we propose a similar optimal transport formulation for video alignment, outperforming previous methods on video alignment datasets, and extend it to a joint video alignment and action segmentation model, establishing the new state of the art on action segmentation benchmarks.

\noindent \textbf{Optimal Transport with Structured Data.}
Optimal transport underpins several computer vision and machine learning applications. A comprehensive review of optimal transport and its applications in machine learning is presented in Khamis et al.~\cite{khamis2024scalable}. Optimal transport applications in computer vision include keypoint matching~\cite{liu2020semantic,sarlin2020superglue}, point set registration~\cite{shen2021accurate}, object detection~\cite{de2023unbalanced}, object tracking~\cite{lee2024sota}, video alignment~\cite{liu2022learning}, and procedure learning~\cite{chowdhuryopel}. For problems with structured data, a Gromov-Wasserstein optimal transport with a structural prior is frequently employed, e.g., graph matching~\cite{xu2019gromov}, brain image registration~\cite{thual2022aligning}, and action segmentation~\cite{xu2024temporally}. In this paper, we develop a Gromov-Wasserstein optimal transport with a structural prior for video alignment and extend it to joint video alignment and action segmentation. To our best knowledge, our work is the first to incorporate video alignment and action segmentation into a unified optimal transport framework.

%% file: Sections/method.tex
\section{Our Approach}
\label{sec:method}

We describe in this section our main contributions, namely a self-supervised video alignment approach (VAOT) in Sec.~\ref{sec:single_task_model} and a joint self-supervised video alignment and action segmentation approach (VASOT) in Sec.~\ref{sec:joint_model}. 

\noindent \textbf{Notations.} 
First of all, $\langle \mathbf{A}, \mathbf{B} \rangle = \sum_{i, j} A_{ij} B_{ij}$ denotes the dot product of $\mathbf{A}, \mathbf{B} \in \mathbb{R}^{n \times m}$, $\mathbf{1}_n \in \mathbb{R}^n$ models a vector of ones, and $[n]= \{1, \dots, n\}$ represents a discrete set of $n$ elements. Next, $\Delta_m \subset \mathbb{R}^m$ denotes the $(m-1)$ dimensional probability simplex, while $\Delta_m^n \subset \mathbb{R}^{m \times n}$ models the Cartesian product space consisting of $n$ such simplexes. Furthermore, $X = \{x_1, \dots, x_N\}$ and $Y = \{y_1, \dots, y_M\}$ denote two input videos of $N$ and $M$ frames respectively. Let $f_{\bm{\theta}}$ with learnable parameters $\bm{\theta}$ be the embedding function, frame-level embeddings of $X$ and $Y$ are expressed as $\mathbf{X} = f_{\bm{\theta}}(X) \in \mathbb{R}^{N \times D}$ and $\mathbf{Y} = f_{\bm{\theta}}(Y) \in \mathbb{R}^{M \times D}$, where $D$ is the embedding vector length. Lastly, $K$ learnable action centroids are expressed by $\mathbf{A} = \left[ \mathbf{a}_1, \dots, \mathbf{a}_K \right] \in \mathbb{R}^{D \times K}$. 

\subsection{Self-Supervised Video Alignment}
\label{sec:single_task_model}

\subsubsection{Optimal Transport with Structured Data}
\label{sec:ot}

\noindent \textbf{Kantorovich Optimal Transport.}
We briefly describe the conventional optimal transport formulation, also known as Kantorovich optiomal transport (KOT)~\cite{thorpe2018introduction}, in the discrete setting. The KOT problem aims to find the minimum-cost coupling $\mathbf{T}^\star$ between histograms $\mathbf{p}\in \Delta_n$ and $\mathbf{q} \in \Delta_m$ with a ground cost $\mathbf{C}\in\mathbb{R}^{n\times m}_+$ and is written as:
\begin{equation}
\label{eq:ot_prob}
   \argmin_{\mathbf{T}\in \mathcal{T}_{\mathbf{p}, \mathbf{q}}}~~~\mathcal{F}_{\text{KOT}}(\mathbf{C}, \mathbf{T}) = \langle \mathbf{C}, \mathbf{T}\rangle,
\end{equation}
with $\mathcal{T}_{\mathbf{p}, \mathbf{q}} = \{\mathbf{T}\in\mathbb{R}_+^{n\times m} \mid \mathbf{T} \mathbf{1}_m = \mathbf{p}, \mathbf{T}^\top \mathbf{1}_n = \mathbf{q} \}$. The coupling $\mathbf{T}$ is regarded as the \emph{soft assignment} between elements in the supports of $\mathbf{p}$ and $\mathbf{q}$, i.e., discrete sets $[n]$ and $[m]$. For video alignment, $\mathbf{T}\in\mathbb{R}_+^{N\times M}$ represents the assignment between frames of $X$ and $Y$.

\noindent \textbf{Gromov-Wasserstein Optimal Transport.}
For histograms defined over incomparable spaces, Gromov-Wasserstein (GW) optimal transport~\cite{peyre2016gromov} is typically employed as:
\begin{equation}
\label{eq:gwot_obj}
    \argmin_{\mathbf{T}\in \mathcal{T}_{\mathbf{p}, \mathbf{q}}}~~~\mathcal{F}_{\text{GW}}(\mathbf{C}^x, \mathbf{C}^y, \mathbf{T}) = \sum_{\substack{i,k \in [n]\\j,l \in [m]}} L(\mathbf{C}_{ik}^x, \mathbf{C}_{jl}^y) \mathbf{T}_{ij}\mathbf{T}_{kl}.
\end{equation}
Here, $(\mathbf{C}^x, \mathbf{p})\in\mathbb{R}^{n\times n}\times \Delta_n$ and $(\mathbf{C}^y, \mathbf{q})\in\mathbb{R}^{m\times m}\times \Delta_m$ represent two (metric, measure) pairs respectively, while distance matrices $\mathbf{C}^x$ and $\mathbf{C}^y$ describe metrics defined over supports $[n]$ and $[m]$ respectively. Note that there is no metric defined \emph{between} supports $[n]$ and $[m]$ in the GW setting. $L:\mathbb{R}\times \mathbb{R} \rightarrow \mathbb{R}$ denotes a cost function minimizing discrepancies between distance matrix elements. We utilize the GW formulation to impose \emph{structural priors} $\mathbf{C}^x$ and $\mathbf{C}^y$ on the transport map for video alignment (i.e., temporal consistency), which we will describe in the next section.

\noindent \textbf{Fused Gromov-Wasserstein Optimal Transport.}
Fused Gromov-Wasserstein (FGW) optimal transport~\cite{titouan2019optimal,vayer2020fused} which merges KOT and GW formulations is often used for problems with known ground cost and structural prior. Let $\alpha \in [0, 1]$, the FGW problem is expressed as:
\begin{multline}
\label{eq:fgwot_obj}
   \argmin_{\mathbf{T}\in \mathcal{T}_{\mathbf{p}, \mathbf{q}}}~~~\mathcal{F}_{\text{FGW}}(\mathbf{C}, \mathbf{C}^x, \mathbf{C}^y, \mathbf{T}) =  (1-\alpha) \mathcal{F}_{\text{KOT}}(\mathbf{C}, \mathbf{T}) +\\ \alpha\mathcal{F}_{\text{GW}}(\mathbf{C}^x, \mathbf{C}^y, \mathbf{T}).
\end{multline}
For video alignment, the KOT objective encourages visual similarity between corresponding frames of $X$ and $Y$, while the GW objective enforces structural properties on the resulting alignment (i.e., temporal consistency).

\noindent \textbf{Balanced Optimal Transport.}
The above optimal transport problems impose \emph{balanced assignment} constraints $\mathbf{T}\in \mathcal{T}_{\mathbf{p}, \mathbf{q}} = \{\mathbf{T}\in\mathbb{R}_+^{n\times m} \mid \mathbf{T} \mathbf{1}_m = \mathbf{p}, \mathbf{T}^\top \mathbf{1}_n = \mathbf{q} \}$. Recent works have relaxed these constraints by replacing (one~\cite{xu2024temporally} or both~\cite{thual2022aligning}) marginal constraints on $\mathbf{T}$ with penalty terms in the objective, yielding (\emph{partial}~\cite{xu2024temporally} or \emph{full}~\cite{thual2022aligning}) \emph{unbalanced assignment} constraints. For video alignment, we adopt the full unbalanced formulation~\cite{thual2022aligning} but the results are worse than those of the balanced formulation, as we will show later in Sec.~\ref{sec:ablation}. This is likely because it is difficult to balance multiple losses (the full unbalanced formulation yields two extra penalty terms) and video alignment is generally more balanced than action segmentation (the number of frames is much larger than the number of actions).

\subsubsection{Video Alignment Optimal Transport}
\label{sec:vaot}

Here, we adapt the above balanced FGW optimal transport in Eq.~\ref{eq:fgwot_obj} for video alignment, yielding our proposed video alignment optimal transport (VAOT). Let us denote $\mathbf{p} = \frac{1}{N}\mathbf{1}_N$ and $\mathbf{q} = \frac{1}{M}\mathbf{1}_M$ as histograms defined over the sets of $N$ frames in $X$ and $M$ frames in $Y$, represented by $[N]$ and $[M]$ respectively. The solution $\mathbf{T}^\star\in\mathbb{R}_+^{N\times M}$ between $[N]$ and $[M]$ represents the soft assignment between frames of $X$ and $Y$. For a frame $x_i$ in $X$, the corresponding frame $y_{j^\star}$ in $Y$ is specified by $j^\star = \argmax_j\mathbf{T}_{ij}^\star$. Below we will discuss our cost matrices $\{\mathbf{C}, \mathbf{C}^x, \mathbf{C}^y\}$ for the FGW problem in Eq.~\ref{eq:fgwot_obj}, deriving the solution $\mathbf{T}^\star$ efficiently, and handling background/redundant frames.

\noindent \textbf{Visual Cue.}
The KOT subproblem in Eq.~\ref{eq:fgwot_obj} includes the cost matrix $\mathbf{C}$ which measures the difference in visual content of $X$ and $Y$ and is defined as $\mathbf{C}_{ij} = 1 - \frac{\mathbf{x}_i^\top \mathbf{y}_j}{\|\mathbf{x}_i\|_2\|\mathbf{y}_j\|_2}$, with frame embeddings $\mathbf{x}_i = f_{\bm{\theta}}(x_i)$ and $\mathbf{y}_j = f_{\bm{\theta}}(y_j)$.

\noindent \textbf{Structural Prior.}
For the GW subproblem in Eq.~\ref{eq:fgwot_obj}, we define $L(a,b) = ab$ and cost matrices $\mathbf{C}^x\in\mathbb{R}_+^{N\times N}$ over frames of $X$ and $\mathbf{C}^y\in\mathbb{R}_+^{M\times M}$ over frames of $Y$ as:
\begin{equation}
\label{eq:cxcy_def}
    \mathbf{C}^x_{ik} =
    \begin{cases}
        \frac{1}{r} & 1 \leq \delta_{ik} \leq Nr \\
        0 & \text{otherwise}
    \end{cases},
    \mathbf{C}^y_{jl} =
    \begin{cases}
        0 & 1 \leq \delta_{jl} \leq Mr \\
        1 & \text{otherwise}
    \end{cases}.
\end{equation}
Here, $\delta_{ik} = |i-k|$, $\delta_{jl} = |j-l|$, and a radius parameter $r \in (0,1]$. The GW component encourages temporal consistency over $\mathbf{T}$. In particular, assigning temporally nearby frames in $X$ ($\delta_{ik} \leq Nr$) to temporally distant frames in $Y$ ($\delta_{jl} > Mr$) incurs a cost ($L(\mathbf{C}^x_{ik},\mathbf{C}^y_{jl}) = \frac{1}{r}$), whereas mapping temporally nearby frames in $X$ ($\delta_{ik} \leq Nr$) to temporally adjacent frames in $Y$ ($\delta_{jl} \leq Mr$) or mapping temporally distant frames in $X$ ($\delta_{ik} > Nr$) to temporally remote frames in $Y$ ($\delta_{jl} > Mr$) incurs no cost ($L(\mathbf{C}^x_{ik},\mathbf{C}^y_{jl}) = 0$). The GW component is capable of handling order variations and repeated actions, as shown in ASOT~\cite{xu2024temporally}.

\noindent \textbf{Fast Numerical Solver for VAOT.}
The GW component in Eq.~\ref{eq:fgwot_obj} can be computed efficiently as $\mathcal{F}_{\text{GW}}(\mathbf{C}^x, \mathbf{C}^y, \mathbf{T}) = \langle \mathbf{C}^x \mathbf{T} \mathbf{C}^y , \mathbf{T} \rangle$ since the cost function $L(a, b) = ab$ can be factorized~\cite{peyre2016gromov}. In addition, by adding an entropy regularization term $-\epsilon H(\mathbf{T})$, with $H(\mathbf{T}) = -\sum_{i,j} T_{ij}\log T_{ij}$ and $\epsilon > 0$, to the FGW formulation in Eq.~\ref{eq:fgwot_obj}, we can obtain the solution $\mathbf{T}^\star$ efficiently via projected mirror descent~\cite{peyre2016gromov}, which can be run on GPUs. Our solver often converges in less than $25$ iterations. By exploiting the sparse structures of $\mathbf{C}^x$ and $\mathbf{C}^y$, each iteration has $O(NM)$ time complexity.

\noindent \textbf{Background/Redundant Frame Handling.}
To tackle background/redundant frames, we follow VAVA~\cite{liu2022learning} to add a \emph{virtual frame} to $X$ and $Y$ so that background/redundant frames are explicitly assigned to it. Specifically, we append an extra row and column to $\mathbf{T}$ and expand other variables accordingly. If the assignment probability of $x_i$ ($i \leq N$) with every $y_j$ ($j \leq M$) is smaller than a threshold parameter $\zeta$, we match $x_i$ with the virtual frame $y_{M+1}$. Similarly, if the assignment probability of $y_j$ ($j \leq M$) with every $x_i$ ($i \leq N$) is smaller than $\zeta$, we match $y_j$ with $x_{N+1}$. Note that virtual frames and their associated frames are excluded from computing the losses. As shown in Sec.~\ref{sec:ablation}, handling background/redundant frames leads to performance gain.

\begin{figure}[t]
	\centering
		\includegraphics[width=1.0\linewidth, trim = 0mm 5mm 85mm 0mm, clip]{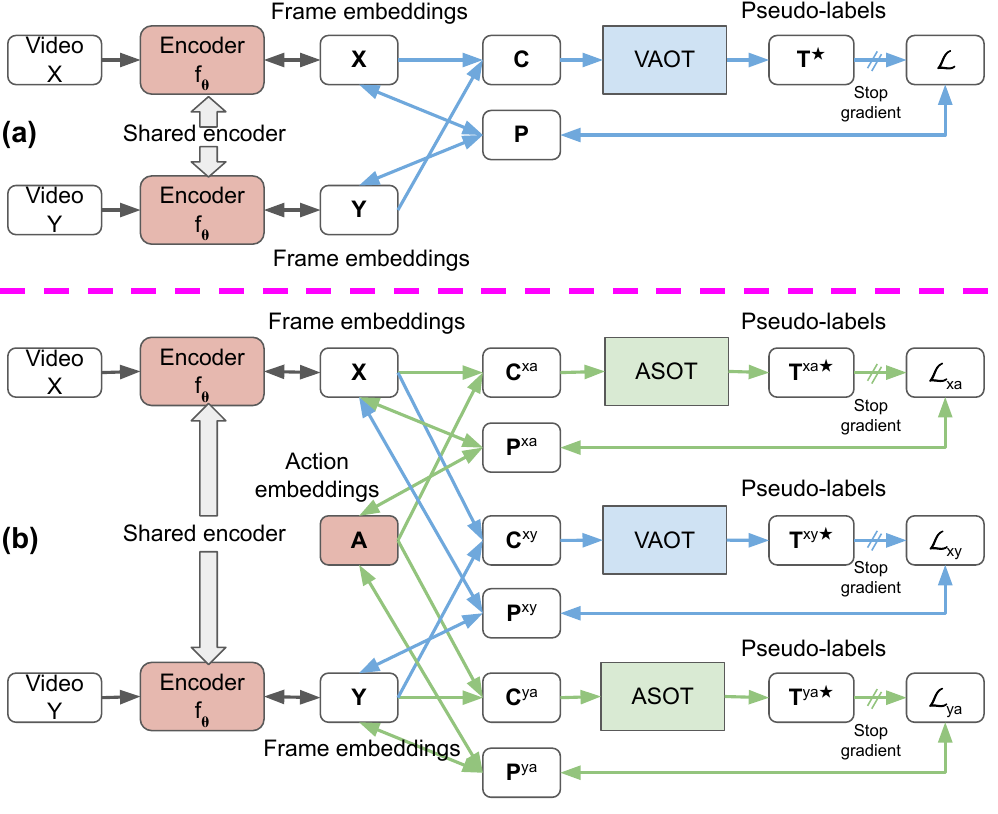}
	\caption{(a) Our self-supervised video alignment method (VAOT). (b) Our joint self-supervised video alignment and action segmentation method (VASOT). Learnable parameters are shown in red. Arrows denote computation/gradient flows (blue and green represent video alignment and action segmentation respectively).}
	\label{fig:method}
\end{figure}

\subsubsection{Self-Supervised Learning}
\label{sec:ssl}

We now present our self-supervised learning framework for video alignment in Fig.~\ref{fig:method}(a). We utilize the above VAOT module to compute pseudo-labels as supervision signals for training the frame encoder. We learn the parameters $\bm{\theta}$ of the frame encoder by minimizing the cross-entropy loss between normalized similarities $\mathbf{P}$ (computed based on frame embeddings $\mathbf{X}$ and $\mathbf{Y}$) and pseudo-labels $\mathbf{T}^\star$ (obtained by VAOT). We first define normalized similarities $\mathbf{P}\in\Delta_M^N$ as: 
\begin{equation}
\label{eq:clus-emb-sims}
    \mathbf{P}_{ij} = \frac{\exp(\mathbf{X} \mathbf{Y}^\top / \tau)_{ij}}{\sum_l\exp(\mathbf{X} \mathbf{Y}^\top / \tau)_{il}},
\end{equation}
with $\tau > 0$ denoting a temperature scaling parameter. In addition, pseudo-labels $\mathbf{T}^\star$ are obtained by solving the balanced FGW problem in Eq.~\ref{eq:fgwot_obj} but with the augmented KOT cost matrix $\Tilde{\mathbf{C}} = \mathbf{C} + \rho \mathbf{R}$, with $\rho \geq 0$. Here, we follow VAVA~\cite{liu2022learning} to impose the \emph{temporal prior} $\mathbf{R}$, defined as $\mathbf{R}_{ij} = \left| i / N - j / M \right|$, which encourages the coupling $\mathbf{T}$ to have a banded diagonal shape and temporally nearby frames in $X$ to be matched with temporally adjacent frames in $Y$, yielding improved performance as seen in Sec.~\ref{sec:ablation}. Finally, our self-supervised learning loss is written as:
\begin{equation}
\label{eq:lxy}
    \mathcal{L} = -\sum_{i=1}^N\sum_{j=1}^M \mathbf{T}^\star_{ij} \log \mathbf{P}_{ij}.
\end{equation}
Note we do not back-propagate gradients through $\mathbf{T}^\star$.

\subsection{Joint Self-Supervised Video Alignment and Action Segmentation}
\label{sec:joint_model}

\subsubsection{Self-Supervised Video Alignment vs. Action Segmentation}
Sec.~\ref{sec:single_task_model} presents our self-supervised video alignment approach (VAOT) developed based on an FGW optimal transport formulation with a structural prior, which has previously been adopted for self-supervised action segmentation by ASOT~\cite{xu2024temporally}. Below we discuss differences between self-supervised video alignment and action segmentation, which lead to distinct design choices for VAOT and ASOT~\cite{xu2024temporally}. Firstly, as illustrated in Fig.~\ref{fig:teaser}, video alignment performs finer-grained \emph{frame-to-frame} assignment, as compared to coarser-grained \emph{frame-to-action} assignment in action segmentation, which causes our removal of $\mathbf{A}$ and our cost matrices $\{\mathbf{C}, \mathbf{C}^x, \mathbf{C}^y\}$ in Sec.~\ref{sec:single_task_model} to be different from ASOT~\cite{xu2024temporally}. Secondly, video alignment generally has a more balanced assignment than action segmentation (the number of frames is much larger than the number of actions) and it is hard to balance multiple losses (the full unbalanced formulation adds two extra penalty terms). Thus, a balanced FGW formulation performs the best for VAOT, whereas a partial unbalanced FGW formulation is preferred in ASOT~\cite{xu2024temporally}. Thirdly, a background action class, to which background/redundant frames are assigned, is typically included in the $K$ actions for action segmentation, whereas it is not already defined for video alignment. Thus, VAOT adds a virtual frame to tackle background/redundant frames. 

\subsubsection{Self-Supervised Multi-Task Learning}
Since both self-supervised video alignment and action segmentation exploit fine-grained temporal information in videos, we propose a self-supervised multi-task learning framework for joint video alignment and action segmentation. In particular, we combine our VAOT module for video alignment with ASOT~\cite{xu2024temporally} for action segmentation into a unified optimal transport-based approach (VASOT), which is illustrated in Fig.~\ref{fig:method}(b). Here, we update variable names in VAOT from \{$\mathbf{C}, \mathbf{P}, \mathbf{T}^\star, \mathcal{L}$\} to \{$\mathbf{C}^{xy}, \mathbf{P}^{xy}, \mathbf{T}^{xy\star}, \mathcal{L}_{xy}$\} respectively for video alignment between $X$ and $Y$, while introducing new variables \{$\mathbf{C}^{xa}, \mathbf{P}^{xa}, \mathbf{T}^{xa\star}, \mathcal{L}_{xa}$\} and \{$\mathbf{C}^{ya}, \mathbf{P}^{ya}, \mathbf{T}^{ya\star}, \mathcal{L}_{ya}$\} for action segmentation on $X$ and $Y$ respectively. The parameters $\bm{\theta}$ of the frame encoder and the action embeddings $\mathbf{A}$ are trained by using the below combination of self-supervised learning losses:
\begin{equation}
\label{eq:vasot}
    \mathcal{L}_{joint} = w_{align} \mathcal{L}_{xy} + w_{seg} (\mathcal{L}_{xa} + \mathcal{L}_{ya}),
\end{equation}
where $w_{align} \geq 0$ and $w_{seg} \geq 0$ denote the weights for the video alignment loss $\mathcal{L}_{xy}$ and the action segmentation losses $\mathcal{L}_{xa}$ and $\mathcal{L}_{ya}$ respectively, $\mathcal{L}_{xy}$ is the cross-entropy loss between normalized similarities $\mathbf{P}^{xy}$ (computed between $\mathbf{X}$ and $\mathbf{Y}$) and pseudo-labels $\mathbf{T}^{xy\star}$ (obtained by VAOT), while $\mathcal{L}_{xa}$ is the cross-entropy loss between normalized similarities $\mathbf{P}^{xa}$ (computed based on $\mathbf{X}$ and $\mathbf{A}$) and pseudo-labels $\mathbf{T}^{xa\star}$ (derived by ASOT~\cite{xu2024temporally}) and $\mathcal{L}_{ya}$ is the cross-entropy loss between normalized similarities $\mathbf{P}^{ya}$ (computed between $\mathbf{Y}$ and $\mathbf{A}$) and pseudo-labels $\mathbf{T}^{ya\star}$ (obtained by ASOT~\cite{xu2024temporally}). This is in contrast with VAOT or ASOT~\cite{xu2024temporally}, where ${\bm{\theta}}$ is solely learned by using either $\mathcal{L}_{xy}$ or $\mathcal{L}_{xa}$ and $\mathcal{L}_{ya}$ respectively. We find balancing $w_{align} = w_{seg} = 1$ yields good results for both video alignment and action segmentation, as seen in Sec.~\ref{sec:sensitivity}. Our joint model requires training and storing a single model, saving both time and memory usage as compared to two single-task models. As we observe in Sec.~\ref{sec:sota_comparison}, in a multi-task learning setting, action segmentation offers little benefit to video alignment performance, while video alignment boosts action segmentation results substantially.

%% file: Sections/experiments.tex
\section{Experiments}
\label{sec:experiments}

\input{Tables/ablation_results}

\noindent \textbf{Datasets.}
We benchmark our VAOT and VASOT approaches for video alignment using \emph{three} datasets, including monotonic datasets, i.e., Pouring~\cite{sermanet2018time} and Penn Action~\cite{zhang2013actemes}, and in-the-wild dataset, i.e., IKEA ASM~\cite{ben2021ikea}. Pouring includes videos of humans pouring liquids and Penn Action comprises of videos of humans playing sports, while IKEA ASM videos capture humans assembling furniture. All methods have the same training and validation splits. Moreover, for action segmentation evaluation, we use \emph{four} datasets, including in-the-wild datasets, i.e., Breakfast~\cite{kuehne2014language}, 50 Salads~\cite{stein2013combining}, and YouTube Instructions~\cite{alayrac2016unsupervised}, and monotonic dataset, i.e., Desktop Assembly~\cite{kumar2022unsupervised}. Breakfast and 50 Salads videos show cooking activities, while YouTube Instructions consists of instructional videos and Desktop Assembly includes videos of an assembly activity. All methods are trained and tested on the same set of videos. For 50 Salads, we evaluate at two action granularity levels, i.e., \emph{Mid} with 19 actions and \emph{Eval} with 12 actions. Finally, for datasets with many activities, i.e., Penn Action, Breakfast, and YouTube Instructions, we train and test the methods per activity and report the average results.

\noindent \textbf{Implementation Details.}
For fair comparison purposes, our VAOT and VASOT approaches for video alignment utilize the same ResNet-50 encoder as recent self-supervised video alignment methods~\cite{dwibedi2019tcc,haresh2021learning,liu2022learning,donahue2024learning}. Similarly, our VASOT approach for action segmentation employs the same MLP encoder as state-of-the-art self-supervised action segmentation methods~\cite{kukleva2019unsupervised,kumar2022unsupervised,xu2024temporally}. Action embeddings $\mathbf{A}$ are initialized via $K$-Means, while the number of clusters $K$ is set to the ground truth value. We implement our methods in PyTorch~\cite{paszke2017automatic} and use ADAM optimization~\cite{kingma2014adam}. Please refer to our supplementary material for more details.

\noindent \textbf{Competing Methods.}
We compare our VAOT and VASOT approaches against prior self-supervised video alignment methods, namely SAL~\cite{misra2016shuffle}, TCN~\cite{sermanet2018time}, TCC~\cite{dwibedi2019tcc}, LAV~\cite{haresh2021learning}, VAVA~\cite{liu2022learning}, and GTCC~\cite{donahue2024learning}. VAVA~\cite{liu2022learning}, which integrates an optimality prior into a classical Kantorovich optimal transport, is the closest to our VAOT approach. Also, we test our VASOT approach against previous self-supervised action segmentation methods, namely CTE~\cite{kukleva2019unsupervised}, VTE~\cite{vidalmata2021joint}, UDE~\cite{swetha2021unsupervised}, ASAL~\cite{li2021action}, TOT~\cite{kumar2022unsupervised}, UFSA~\cite{tran2024permutation}, ASOT~\cite{xu2024temporally}, and HVQ~\cite{spurio2024hierarchical}. ASOT~\cite{xu2024temporally} is the single-task baseline for action segmentation, which we adopt in Sec.~\ref{sec:joint_model} for our multi-task VASOT approach.

\noindent \textbf{Evaluation Metrics.}
To evaluate our VAOT and VASOT approaches for video alignment, we compute \emph{four} metrics on the validation set, i.e., phase classification (\emph{Acc@\{0.1,0.5,1.0\}}), phase progression (\emph{Progress}), video alignment (\emph{$\tau$}), and fine-grained frame retrieval (\emph{AP@\{5,10,15\}}). Prior to that, we train the model on the training set and freeze it, and then train an SVM classifier or linear regressor on top of frozen features. Note that as mentioned in~\cite{dwibedi2019tcc,haresh2021learning,liu2022learning,donahue2024learning}, Progress and $\tau$ are only defined for monotonic datasets and hence are not computed for IKEA ASM. Also, to test our VASOT approach for action segmentation, we calculate \emph{three} metrics, i.e., mean over frames (\emph{MoF}), F1 score (\emph{F1}), and mean intersection over union (\emph{mIoU}). Before that, we train the model, obtain predicted action segments, and perform Hungarian matching between predicted and ground truth action clusters.

\subsection{Ablation Analysis Results}
\label{sec:ablation}

We first study the impacts of design choices in VAOT in Sec.~\ref{sec:single_task_model}. We show the IKEA ASM results in Tab.~\ref{tab:ablation_results}. Please see our supplementary material for the Pouring results.

\noindent \textbf{Effect of Structural Prior.}
The structural priors $\{\mathbf{C}^x, \mathbf{C}^y\}$ defined in Eq.~\ref{eq:cxcy_def} are used to encourage temporal consistency on the transport map $\mathbf{T}$. We analyze the impact of the structural priors by removing the GW subproblem in Eq.~\ref{eq:fgwot_obj} (via setting $\alpha = 0$), yields worse results, as reported in Tab.~\ref{tab:ablation_results}. This demonstrates the importance of the structural priors and temporal consistency in our VAOT approach.

\noindent \textbf{Effect of Temporal Prior.} 
Removing the temporal prior $\mathbf{R}$ described in Sec.~\ref{sec:ssl} (by setting $\rho = 0$), leading to notable performance drops in Tab.~\ref{tab:ablation_results}. This validates the contribution of the temporal prior in our VAOT approach.

\noindent \textbf{Effect of Balanced Assignment.}
When the balanced assignment formulation is replaced by the full unbalanced assignment formulation, the performance degrades significantly, as shown in Tab.~\ref{tab:ablation_results}. This indicates that the balanced assignment formulation is preferred for the video alignment problem, as we discussed previously in Sec.~\ref{sec:ot}. 

\noindent \textbf{Effect of Virtual Frame.}
Virtual frames (described in Sec.~\ref{sec:vaot}) are used to tackle background/redundant frames. From Tab.~\ref{tab:ablation_results}, removing virtual frames negatively affects the robustness and hence performance of our VAOT approach. 

\subsection{Sensitivity Analysis Results}
\label{sec:sensitivity}

\begin{figure}[t]
     \centering
     \begin{subfigure}[t]{0.23\textwidth}
         \centering
         \includegraphics[width=\textwidth]{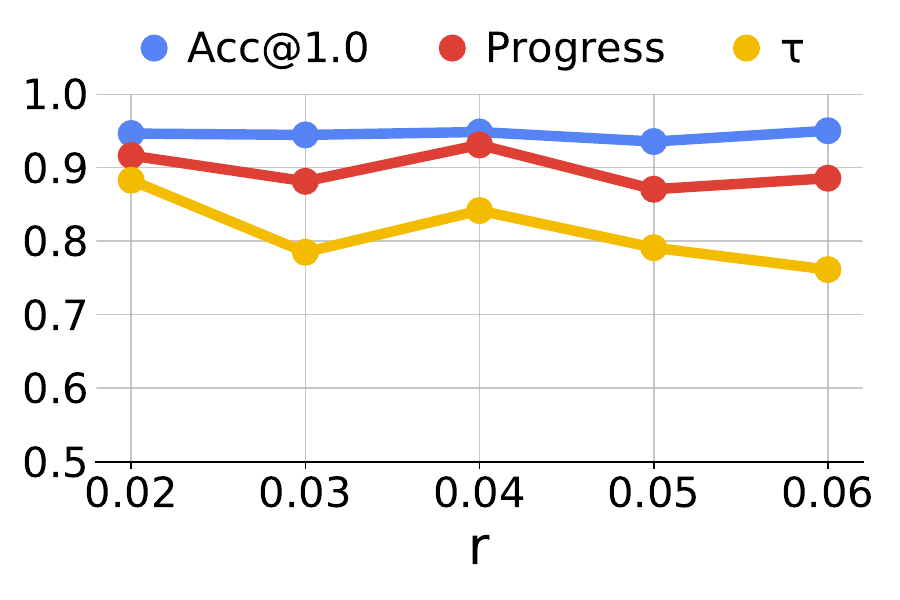}
         \caption{Structural prior radius $r$.}
         \label{fig:sens_r}
     \end{subfigure}     
     \begin{subfigure}[t]{0.23\textwidth}
         \centering
         \includegraphics[width=\textwidth]{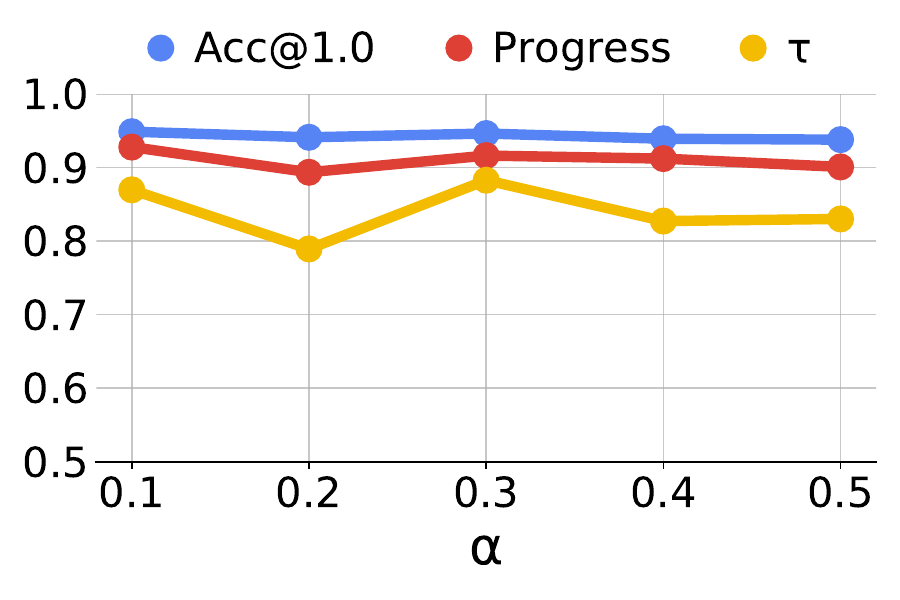}
         \caption{Gromov-Wasserstein weight $\alpha$.}
         \label{fig:sens_alpha}
     \end{subfigure}
     \begin{subfigure}[t]{0.23\textwidth}
         \centering
         \includegraphics[width=\textwidth]{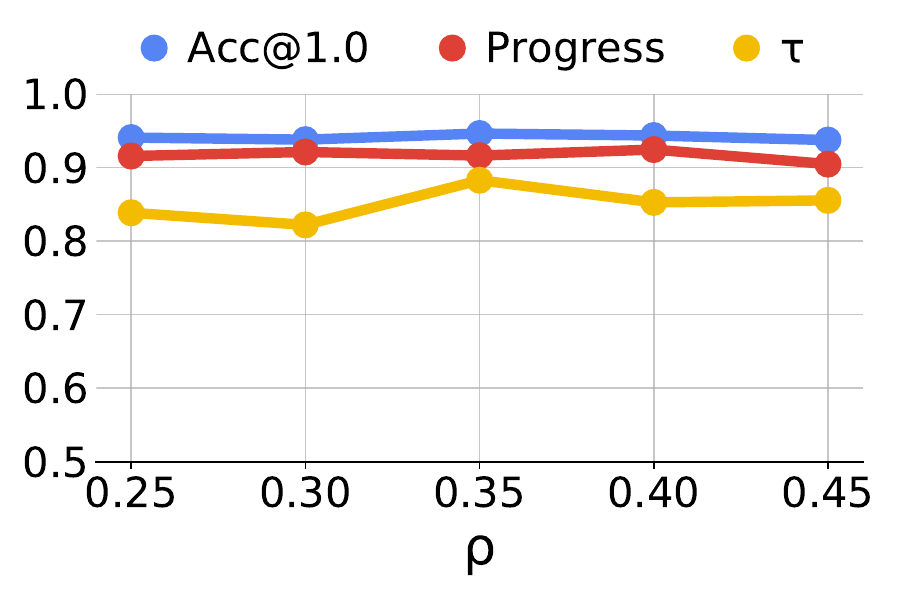}
         \caption{Temporal prior weight $\rho$.}
         \label{fig:sens_rho}
     \end{subfigure}
      \begin{subfigure}[t]{0.23\textwidth}
         \centering
         \includegraphics[width=\textwidth]{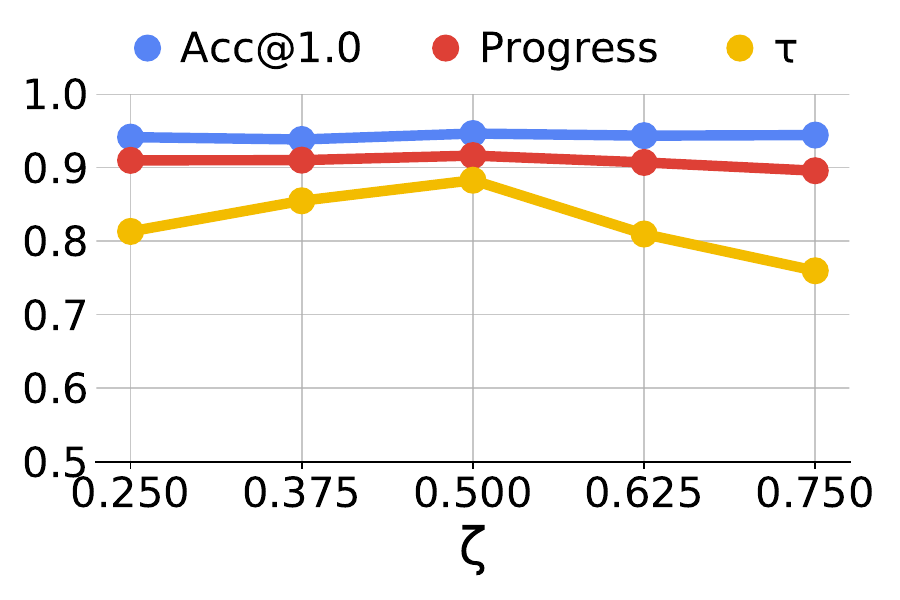}
         \caption{Virtual frame threshold $\zeta$.}
         \label{fig:sens_zeta}
     \end{subfigure}
     \hrule
     \begin{subfigure}[t]{0.23\textwidth}
         \centering
         \includegraphics[width=\textwidth]{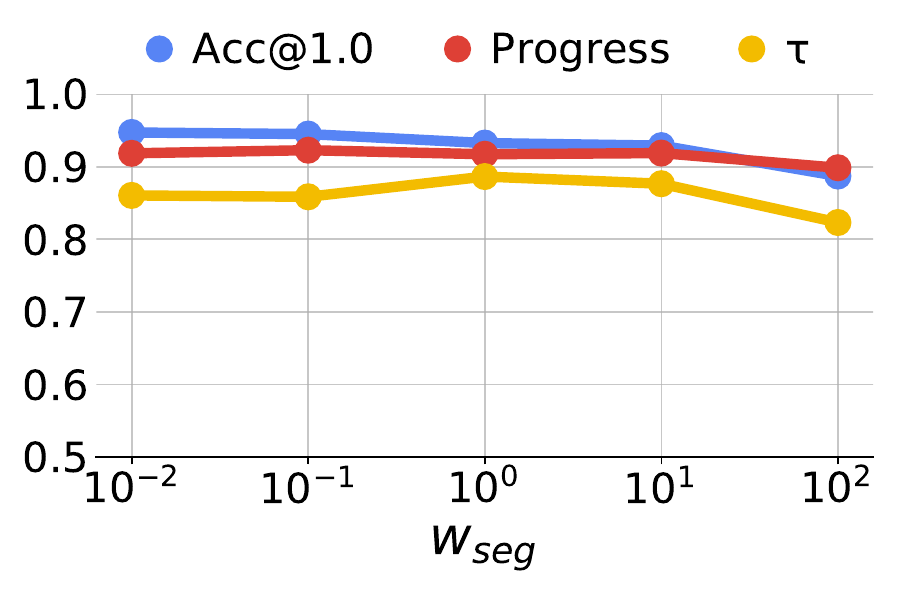}
         \caption{Segmentation loss weight $w_{seg}$ (here, $w_{align} = 1$).}
         \label{fig:sens_wseg}
     \end{subfigure}
     \begin{subfigure}[t]{0.23\textwidth}
         \centering
         \includegraphics[width=\textwidth]{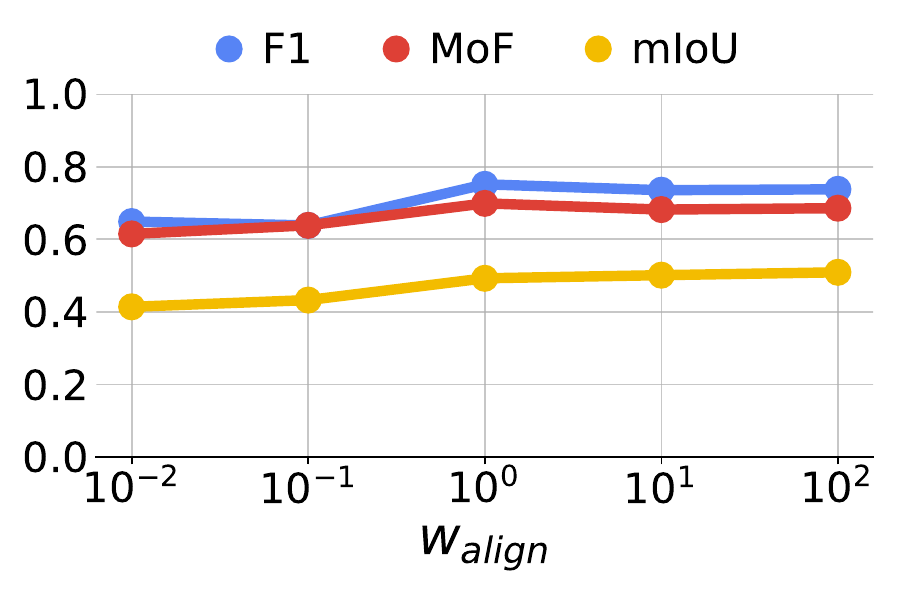}
         \caption{Alignment loss weight $w_{align}$ (here, $w_{seg} = 1$).}
         \label{fig:sens_walign}
     \end{subfigure}
    \caption{Sensitivity analysis results. Note that (a-d) are for VAOT, while (e-f) are for VASOT.}
    \label{fig:sensitivity}
\end{figure}

Here, we conduct sensitivity analyses on hyperparameters of VAOT and VASOT. Fig.~\ref{fig:sensitivity} shows the results. We use Pouring in Figs.~\ref{fig:sensitivity}(a-e) and Desktop Assembly in Fig.~\ref{fig:sensitivity}(f). The results of $\epsilon$ are provided in our supplementary material.

\noindent \textbf{Effect of $r$ and $\alpha$.}
From Fig.~\ref{fig:sensitivity}(a), Acc@1.0 remains stable and Progress shows small variation across all studied values of $r$, whereas $\tau$ is the most sensitive metric, peaking at $r = 0.02$ and decreasing as $r$ increases. Similar observations can be made for $\alpha$ in Fig.~\ref{fig:sensitivity}(b), where Acc@1.0 and Progress are mostly stable, whereas $\tau$ fluctuates the most, performing the best with $\alpha = 0.3$. Furthermore, we find that $r = 0.02$ and $\alpha = 0.3$ also work the best for the remaining datasets.

\noindent \textbf{Effect of $\rho$ and $\zeta$.}
For $\rho$ in Fig.~\ref{fig:sensitivity}(c), it can be seen that Acc@1.0 and Progress remain mostly stable, whereas $\tau$ varies the most, performing the best at $\rho = 0.35$. Similarly, for $\zeta$ in Fig.~\ref{fig:sensitivity}(d), we observe that $\tau$ is the most sensitive metric, peaking at $\zeta = 0.5$, whereas Acc@1.0 and Progress remain steady across the analyzed value range of $\zeta$. Moreover, we notice that $\rho = 0.35$ and $\zeta = 0.5$ also yield the best results for the remaining datasets.

\noindent \textbf{Effect of $w_{seg}$ and $w_{align}$.}
We study the relationship between action segmentation and video alignment in our multi-task VASOT approach by varying their weights $w_{seg}$ and $w_{align}$ in Eq.~\ref{eq:vasot}. In particular, we report alignment results with various $w_{seg}$ values (while keeping $w_{align} = 1$) in Fig.~\ref{fig:sensitivity}(e), and segmentation results with various $w_{align}$ values (while keeping $w_{seg} = 1$) in Fig.~\ref{fig:sensitivity}(f). It can be seen from Fig.~\ref{fig:sensitivity}(e) that alignment results drop as $w_{seg}$ increases and the best overall alignment performance is obtained with $w_{seg} = 1$. In contrast, it is clear from Fig.~\ref{fig:sensitivity}(f) that segmentation results improve as $w_{align}$ increases and become steady after $w_{align}$ reaches $1$. Therefore, in a multi-task learning setup, action segmentation provides little boost to video alignment results, whereas video alignment increases action segmentation performance notably. Moreover, with balancing $w_{seg} = w_{align} = 1$, VASOT achieves good results for both video alignment and action segmentation. For the next section, we set $w_{seg} = w_{align} = 1$ for VASOT.

\subsection{State-of-the-Art Comparison Results}
\label{sec:sota_comparison}

\input{Tables/alignment_results}
\input{Tables/segmentation_results}

\begin{figure}[t]
    \centering
    \includegraphics[width=0.45\textwidth, trim = 5mm 25mm 5mm 26mm, clip]{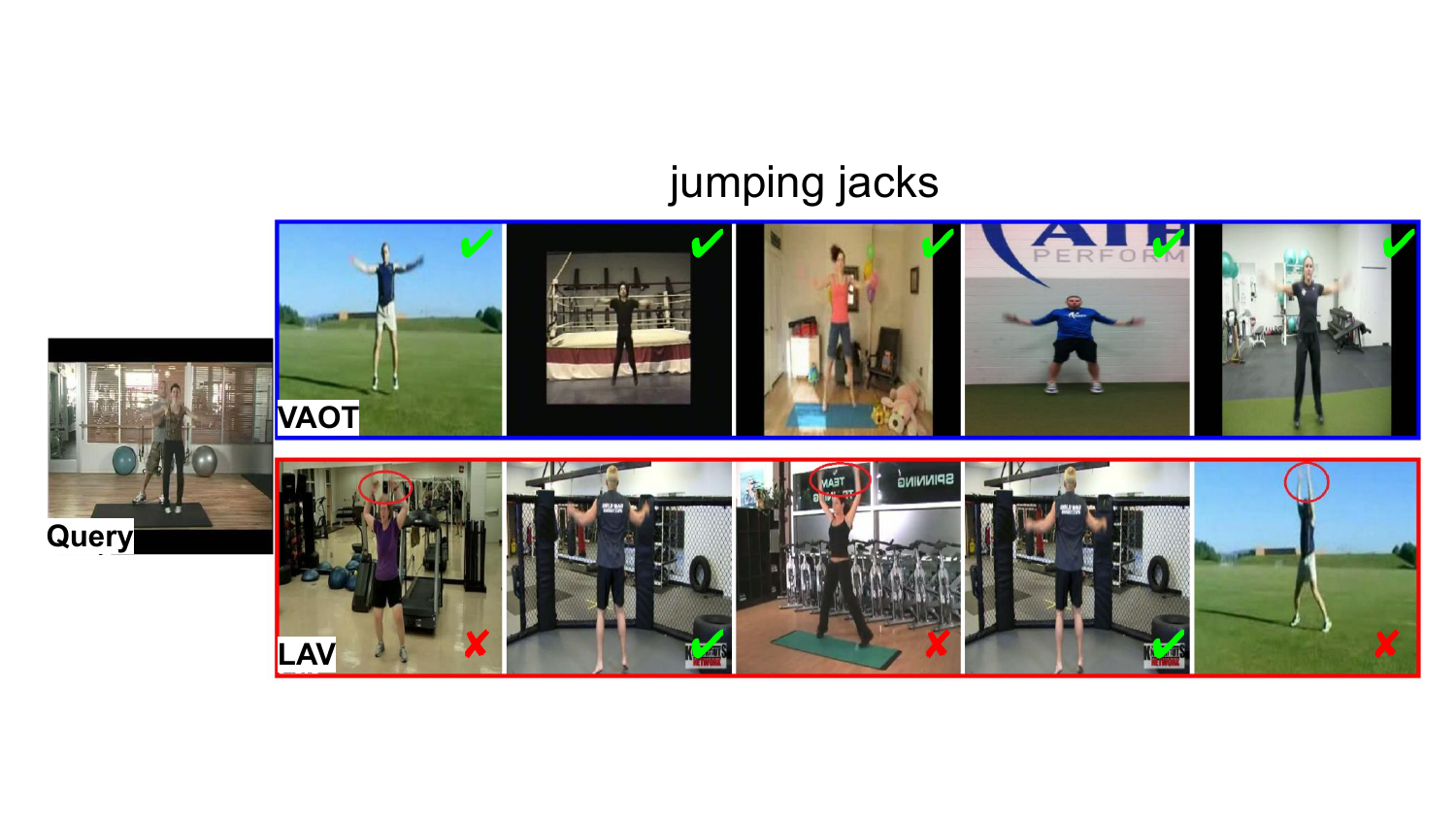} 
    \caption{Fine-grained frame retrieval results on Penn Action. The query image is on the left, while on the right are the top 5 matching images retrieved by VAOT (blue box) and LAV (red box).}
    \label{fig:frame_retrieval}
\end{figure}

\begin{figure}[t]
    \centering
    \includegraphics[width=0.38\textwidth, trim = 0mm 10mm 10mm 0mm, clip]{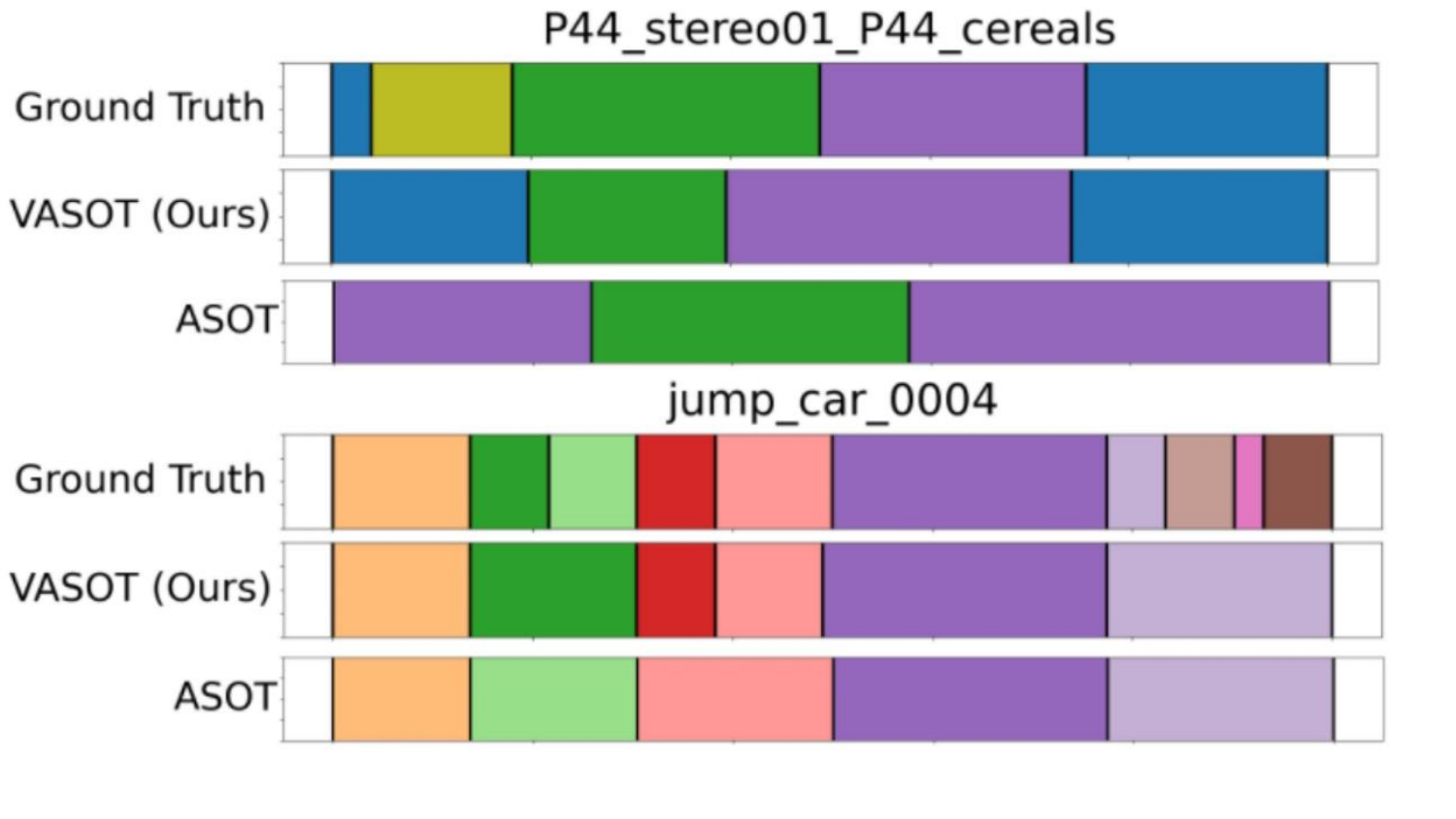} 
    \caption{Action segmentation results on Breakfast (top) and YouTube Instructions (bottom).}
    \label{fig:segmentation}
\end{figure}

\noindent \textbf{Video Alignment Comparison Results.}
We now benchmark our VAOT and VASOT approaches against previous self-supervised video alignment methods and present quantitative results in Tab.~\ref{tab:alignment_results}. Firstly, it is evident from Tab.~\ref{tab:alignment_results} that our VAOT approach achieves the best overall performance across all datasets, outperforming all competing methods, i.e., SAL~\cite{misra2016shuffle}, TCN~\cite{sermanet2018time}, TCC~\cite{dwibedi2019tcc}, LAV~\cite{haresh2021learning}, VAVA~\cite{liu2022learning}, and GTCC~\cite{donahue2024learning}. Especially, VAOT shows major improvements over VAVA~\cite{liu2022learning} on the in-the-wild IKEA ASM dataset. The results confirm the advantage of our FGW formulation with a structural prior over the classical Kantorovich formulation with an optimality prior in VAVA~\cite{liu2022learning}. Next, Fig.~\ref{fig:frame_retrieval} shows some qualitative results, where VAOT retrieves all 5 correct frames with the same action (\emph{Hands at shoulder}) as the query image, while LAV~\cite{haresh2021learning} obtains 3 incorrect frames (\emph{Hands above head}), highlighted by red ovals. Lastly, similar to Fig.~\ref{fig:sensitivity}(e), we find that action segmentation offers little benefit to video alignment in a multi-task learning setup, and our multi-task VASOT approach performs mostly similarly to our single-task VAOT approach in Tab.~\ref{tab:alignment_results}. This is likely because video alignment is a more complex problem involving finer-grained \emph{frame-to-frame} assignment, as compared to coarser-grained \emph{frame-to-action} assignment in action segmentation. Nevertheless, VASOT obtains mostly favorable results over prior works.

\noindent \textbf{Action Segmentation Comparison Results.}
We test our VASOT approach against state-of-the-art self-supervised action segmentation methods and include quantitative results in Tab.~\ref{tab:segmentation_results}. Firstly, from Tab.~\ref{tab:segmentation_results}, VASOT consistently achieves the best results across all metrics and datasets, outperforming all competing methods, i.e., CTE~\cite{kukleva2019unsupervised}, VTE~\cite{vidalmata2021joint}, UDE~\cite{swetha2021unsupervised}, ASAL~\cite{li2021action}, TOT~\cite{kumar2022unsupervised}, UFSA~\cite{tran2024permutation}, ASOT~\cite{xu2024temporally}, and HVQ~\cite{spurio2024hierarchical}. While our multi-task VASOT approach shows small gains over the single-task ASOT baseline~\cite{xu2024temporally} on Breakfast and YouTube Instructions, our improvements on 50 Salads and Desktop Assembly are substantial. The results validate the benefit of fusing video alignment with action segmentation and demonstrate that video alignment boosts action segmentation results notably in a multi-task learning setup. Moreover, Fig.~\ref{fig:segmentation} shows some qualitative results, where VASOT predicts segmentations which capture action boundaries more accurately and are more closely aligned with ground truth than ASOT~\cite{xu2024temporally}.

%% file: Tables/ablation_results.tex
\begin{table*}[t]
\begin{minipage}{\linewidth}
\centering
\footnotesize
{
\begin{tabular}{c|c|c|c|c|c|c|c|c|c}
\specialrule{1pt}{1pt}{1pt}
& \textbf{Method} & \textbf{Acc@0.1} & \textbf{Acc@0.5} & \textbf{Acc@1.0} & \textbf{Progress} & \bm{$\tau$} & \textbf{AP@5} & \textbf{AP@10} & \textbf{AP@15} \\

\midrule
\multirow{5}{*}{\rotatebox[origin=c]{90}{\textbf{IKEA ASM}}}
&w/o Structural Prior
& \underline{30.29} & \underline{35.52} & \underline{37.81} & - & - & 27.54 & 27.33 & 27.15 \\
&w/o Temporal Prior
& 17.84 & 17.84 & 17.84 & - & - & 15.63 & 15.64 & 15.56 \\
&w/o Balanced Assignment
& 17.84 & 20.71 & 25.24 & - & - & 15.49 & 15.69 & 15.78 \\
&w/o Virtual Frame
& 30.16 & 34.49 & 36.10 & - & - & \underline{29.57} & \underline{29.24} & \underline{28.87} \\
&All
& \textbf{33.73} & \textbf{36.42} & \textbf{38.64} & - & - & \textbf{31.49} & \textbf{31.92} & \textbf{32.01} \\

\specialrule{1pt}{1pt}{1pt}
\end{tabular}
}
\caption{Ablation analysis results. \textbf{Bold} and \underline{underline} denote the best and second best respectively.}
\label{tab:ablation_results}
\end{minipage}
\end{table*}

%% file: Tables/alignment_results.tex
\begin{table*}[t]
\begin{minipage}{\linewidth}
\centering
\footnotesize
{
\begin{tabular}{c|c|c|c|c|c|c|c|c|c}
\specialrule{1pt}{1pt}{1pt}
& \textbf{Method} & \textbf{Acc@0.1} & \textbf{Acc@0.5} & \textbf{Acc@1.0} & \textbf{Progress} & \bm{$\tau$} & \textbf{AP@5} & \textbf{AP@10} & \textbf{AP@15} \\

\midrule
\multirow{8}{*}{\rotatebox[origin=c]{90}{\textbf{Pouring}}}
&SAL~\cite{misra2016shuffle}
& 85.68 & 87.84 & 88.02 & 74.51 & 73.31 & 84.05 & 83.77 & 83.79\\
&TCN~\cite{sermanet2018time}
& 89.19 & 90.39 & 90.35 & 80.57 & 86.69 & 83.56 & 83.31 & 83.01\\
&TCC~\cite{dwibedi2019tcc}
& 89.23 & 91.43 & 91.82 & 80.30 & 85.16 & 87.16 & 86.68 & 86.54\\
&LAV~\cite{haresh2021learning}
& 91.61 & \underline{92.82} & 92.84 & 80.54 & 85.61 & 89.13 & 89.13 & 89.22\\
&VAVA~\cite{liu2022learning}
& 91.65 & 91.79 & 92.45 & 83.61 & 87.55 & 90.05 & 89.92 & 90.17\\
&GTCC~\cite{donahue2024learning}
& 71.20 & 89.20 & \underline{93.50} & 85.80 & 88.10 & - & - & -\\
&\cellcolor{beaublue} VAOT (Ours)
&\cellcolor{beaublue} \underline{91.80} &\cellcolor{beaublue} \textbf{92.88} &\cellcolor{beaublue} \textbf{94.63} &\cellcolor{beaublue} \underline{91.63} &\cellcolor{beaublue} \underline{88.28} &\cellcolor{beaublue} \textbf{91.34} &\cellcolor{beaublue} \textbf{90.56} &\cellcolor{beaublue} \textbf{90.29}\\
&\cellcolor{beaublue} VASOT (Ours)
&\cellcolor{beaublue} \textbf{91.93} &\cellcolor{beaublue} 92.12 &\cellcolor{beaublue} 93.04 &\cellcolor{beaublue} \textbf{91.71} &\cellcolor{beaublue} \textbf{88.64} &\cellcolor{beaublue} \underline{91.03} &\cellcolor{beaublue} \underline{90.44} &\cellcolor{beaublue} \underline{90.21}\\

\midrule
\multirow{8}{*}{\rotatebox[origin=c]{90}{\textbf{Penn Action}}}
&SAL~\cite{misra2016shuffle}
& 74.87 & 78.26 & 79.96 & 59.43 & 63.36 & 76.04 & 75.77 & 75.61\\
&TCN~\cite{sermanet2018time}
& 81.99 & 83.67 & 84.04 & 67.62 & 73.28 & 77.84 & 77.51 & 77.28\\
&TCC~\cite{dwibedi2019tcc}
& 81.26 & 83.35 & 84.45 & 67.26 & 73.53 & 76.74 & 76.27 & 75.88\\
&LAV~\cite{haresh2021learning}
& 83.56 & 83.95 & 84.25 & 66.13 & 80.47 & 79.13 & 78.98 & 78.90 \\
&VAVA~\cite{liu2022learning}
& 83.89 & 84.23 & 84.48 & 70.91 & 80.53 & \underline{81.52} & \underline{80.47} & \underline{80.67}\\
&GTCC~\cite{donahue2024learning}
& 78.30 & 81.20 & 81.30 & 70.80 & 88.30 & - & - & - \\
&\cellcolor{beaublue} VAOT (Ours)
&\cellcolor{beaublue} \underline{83.96} &\cellcolor{beaublue} \underline{85.35} &\cellcolor{beaublue} \underline{86.92} &\cellcolor{beaublue} \textbf{84.31} &\cellcolor{beaublue} \textbf{88.99} &\cellcolor{beaublue} \textbf{81.62} &\cellcolor{beaublue} \textbf{81.03} &\cellcolor{beaublue} \textbf{80.68}\\
&\cellcolor{beaublue} VASOT (Ours)
&\cellcolor{beaublue} \textbf{84.17} &\cellcolor{beaublue} \textbf{85.54} &\cellcolor{beaublue} \textbf{87.66} &\cellcolor{beaublue} \underline{83.39} &\cellcolor{beaublue} \underline{88.70} &\cellcolor{beaublue} 78.85 &\cellcolor{beaublue} 78.38 &\cellcolor{beaublue} 78.03\\

\midrule
\multirow{7}{*}{\rotatebox[origin=c]{90}{\textbf{IKEA ASM}}}
&SAL~\cite{misra2016shuffle}
& 22.94 & 23.43 & 25.46 & - & - & 14.28 & 14.04 & 14.10\\
&TCN~\cite{sermanet2018time}
& 22.51 & 25.47 & 25.88 & - & - & 17.37 & 17.03 & 16.96\\
&TCC~\cite{dwibedi2019tcc}
& 22.70 & 25.04 & 25.63 & - & - & 18.03 & 17.53 & 17.20\\
&LAV~\cite{haresh2021learning}
& 23.19 & 25.47 & 25.54 & - & - & 20.14 & 19.35 & 19.21\\
&VAVA~\cite{liu2022learning}
& 29.12 & 29.95 & 29.10 & - & - & 26.42 & 25.73 & 25.80\\
&\cellcolor{beaublue} VAOT (Ours)
&\cellcolor{beaublue} \textbf{33.73} &\cellcolor{beaublue} \textbf{36.42} &\cellcolor{beaublue} \textbf{38.64} &\cellcolor{beaublue} - &\cellcolor{beaublue} - &\cellcolor{beaublue} \textbf{31.49} &\cellcolor{beaublue} \textbf{31.92} &\cellcolor{beaublue} \textbf{32.01}\\
&\cellcolor{beaublue} VASOT (Ours)
&\cellcolor{beaublue} \underline{29.96} &\cellcolor{beaublue} \underline{30.78} &\cellcolor{beaublue} \underline{31.02} &\cellcolor{beaublue} - &\cellcolor{beaublue} - &\cellcolor{beaublue} \underline{30.29} &\cellcolor{beaublue} \underline{30.37} &\cellcolor{beaublue} \underline{30.42}\\

\specialrule{1pt}{1pt}{1pt}
\end{tabular}
}
\caption{Video alignment comparison results. \textbf{Bold} and \underline{underline} denote the best and second best respectively.}
\label{tab:alignment_results}
\end{minipage}
\end{table*}

%% file: Tables/segmentation_results.tex
\begin{table*}[t]
\centering
\footnotesize
\begin{tabular}{c|c|c|c|c|c|c}
\toprule
& \multirow{2}{*}{\textbf{Method}} & \textbf{Breakfast} & \textbf{YouTube Instructions} & \textbf{50 Salads (Mid)} & \textbf{50 Salads (Eval)} & \textbf{Desktop Assembly} \\
\cmidrule(lr){3-3} \cmidrule(lr){4-4} \cmidrule(lr){5-5} \cmidrule(lr){6-6} \cmidrule(lr){7-7}
& & \textbf{MoF / F1 / mIoU} & \textbf{MoF / F1 / mIoU} & \textbf{MoF / F1 / mIoU} & \textbf{MoF / F1 / mIoU} & \textbf{MoF / F1 / mIoU} \\

\midrule
\multirow{11}{*}{\rotatebox[origin=c]{90}{\textbf{Full-Dataset Evaluation}}} & CTE*~\cite{kukleva2019unsupervised} & 41.8 / 26.4 / - & 39.0 / 28.3 / - & 30.2 / - / - & 35.5 / - / - & 47.6 / 44.9 / - \\
& CTE\textdagger~\cite{kukleva2019unsupervised} & 47.2 / 27.0 / 14.9 & 35.9 / 28.0 / 9.9 & 30.1 / 25.5 / 17.9 & 35.0 / 35.5 / 21.6 & - / - / - \\
& VTE~\cite{vidalmata2021joint} & 48.1 / - / - & - / 29.9 / - & 24.2 / - / - & 30.6 / - / - & - / - / - \\
& UDE~\cite{swetha2021unsupervised} & 47.4 / 31.9 / - & 43.8 / 29.6 / - & - / - / - & 42.2 / 34.4 / - & - / - / - \\
& ASAL~\cite{li2021action} & 52.5 / 37.9 / - & 44.9 / 32.1 / - & 34.4 / - / - & 39.2 / - / - & - / - / - \\
& TOT~\cite{kumar2022unsupervised} & 47.5 / 31.0 / - & 40.6 / 30.0 / - & 31.8 / - / - & 47.4 / 42.8 / - & 56.3 / 51.7 / - \\
& TOT+~\cite{kumar2022unsupervised} & 39.0 / 30.3 / - & 45.3 / 32.9 / - & 34.3 / - / - & 44.5 / 48.2 / - & 58.1 / 53.4 / - \\
& UFSA (M)~\cite{tran2024permutation} & - / - / - & 43.2 / 30.5 / - & - / - / - & 47.8 / 34.8 / - & - / - / - \\
& UFSA (T)~\cite{tran2024permutation} & 52.1 / 38.0 / - & 49.6 / 32.4 / - & 36.7 / 30.4 / - & 55.8 / 50.3 / - & 65.4 / 63.0 / - \\
& ASOT~\cite{xu2024temporally} & \underline{56.1} / 38.3 / \underline{18.6} & \underline{52.9} / \underline{35.1} / \underline{24.7} & \underline{46.2} / \underline{37.4} / \underline{24.9} & \underline{59.3} / \underline{53.6} / \underline{30.1} & \underline{70.4} / \underline{68.0} / \underline{45.9} \\
& HVQ~\cite{spurio2024hierarchical} & 54.4 / \textbf{39.7} / - & 50.3 / \underline{35.1} / - & - / - / - & - / - / - & - / - / - \\
&\cellcolor{beaublue} VASOT (Ours) &\cellcolor{beaublue} \textbf{57.5} / \underline{39.0} / \textbf{18.8} &\cellcolor{beaublue} \textbf{53.2} / \textbf{35.7} / \textbf{25.2} &\cellcolor{beaublue} \textbf{47.2} / \textbf{41.3} / \textbf{26.1} &\cellcolor{beaublue} \textbf{60.6} / \textbf{57.4} / \textbf{34.5} &\cellcolor{beaublue} \textbf{70.9} / \textbf{75.1} / \textbf{49.3} \\
        
\bottomrule
\end{tabular}
\caption{Action segmentation comparison results. \textbf{Bold} and \underline{underline} denote the best and second best respectively.}
\label{tab:segmentation_results}
\end{table*}

%% file: Sections/conclusion.tex
\section{Conclusion}
\label{sec:conclusion}

This paper presents a novel approach for joint self-supervised video alignment and action segmentation. We first develop a fused Gromov-Wasserstein optimal transport with a structural prior for self-supervised video alignment, outperforming prior works. Our single-task method trains efficiently on GPUs and needs few iterations to derive the optimal transport solution. Next, we extend our approach to a unified optimal transport framework for joint self-supervised video alignment and action segmentation, yielding similar video alignment yet better action segmentation results than prior works. Our multi-task method requires training and storing a single model and saves both time and memory usage. To our best knowledge, our work is the first to explore the relationship between video alignment and action segmentation. Our future works will explore deep supervision~\cite{li2017deep,li2018deep}, complex weighting in  multi-task learning~\cite{kendall2018multi,chen2018gradnorm}, and other potential applications (e.g., joint keypoint matching and clustering~\cite{estrada2009appearance,sarlin2020superglue}).

%% file: supp.tex
\section{Supplementary Material}
In this supplementary material, we first provide the implementation details of our VAOT and VASOT approaches in Sec.~\ref{sec:supp_implementation}. Next, our ablation analysis results on the Pouring dataset are included in Sec.~\ref{sec:supp_ablation}, while our sensitivity analysis results of the entropy regularization weight $\epsilon$ are presented in Sec.~\ref{sec:supp_sensitivity}. Furthermore, our multi-action video alignment results and our per-video action segmentation results are included in Secs.~\ref{sec:supp_multi_action} and~\ref{sec:supp_per_video} respectively. Finally, we present some qualitative results in Sec.~\ref{sec:supp_qualitative} and complexity comparison results in Sec.~\ref{sec:supp_complexity}.

\subsection{Implementation Details}
\label{sec:supp_implementation}

\input{Tables/supp_alignment_hyperparams}
\input{Tables/supp_seg_hyperparameters}

For fair comparisons with previous self-supervised video alignment methods~\cite{dwibedi2019tcc,haresh2021learning,liu2022learning,donahue2024learning}, our VAOT and VASOT approaches for video alignment utilize a ResNet-50 encoder network. Please refer to Dwibedi et al.~\cite{dwibedi2019tcc} for additional details on the encode network. We provide the hyperparameter settings of our VAOT and VASOT approaches for video alignment in Tab.~\ref{tab:hyperparams}. Note that our VASOT approach for video alignment includes both VAOT and ASOT~\cite{xu2024temporally} modules, which share the same hyperparameter settings shown in the VASOT column in Tab.~\ref{tab:hyperparams}.

In addition, following state-of-the-art self-supervised action segmentation methods~\cite{kukleva2019unsupervised,kumar2022unsupervised,xu2024temporally}, our VASOT approach for action segmentation employs a 2-layer MLP encoder network for fair comparison purposes. Please see Kukleva et al.~\cite{kukleva2019unsupervised} for more details on the encoder network. The hyperparameter settings of our VASOT approach for action segmentation are included in Tab.~\ref{tab:seg_hyperparams}. Note that both VAOT and ASOT~\cite{xu2024temporally} components in our VASOT approach for action segmentation have the same hyperparameter settings presented in the VASOT column in Tab.~\ref{tab:seg_hyperparams}.

\subsection{Ablation Analysis Results}
\label{sec:supp_ablation}

\input{Tables/supp_ablation_results}

In addition to the results on IKEA ASM in Sec.~4.1 of the main paper, we present the ablation analysis results of our VAOT approach on the Pouring dataset in Tab.~\ref{tab:pouring_ablation_results}.

\noindent \textbf{Effect of Structural Prior.}  
The structural prior imposes temporal consistency on the transport map. Removing it significantly degrades performance across all metrics, highlighting its critical role in our VAOT approach.

\noindent \textbf{Effect of Temporal Prior.}  
Excluding the temporal prior results in performance drops across all metrics, confirming its positive impact on video alignment.

\noindent \textbf{Effect of Balanced Assignment.}  
Using a full unbalanced assignment formulation causes a substantial decrease in performance. This demonstrates the importance of balanced assignment in our VAOT approach.

\noindent \textbf{Effect of Virtual Frame.}  
Virtual frames are added for tackling background/redundant frames and improving robustness. Removing virtual frames results in minor performance drops across all metrics, which is expected given the monotonic nature of the Pouring dataset.

\subsection{Sensitivity Analysis Results}
\label{sec:supp_sensitivity}

\begin{figure}[t]
     \centering
     \includegraphics[width=0.48\textwidth]{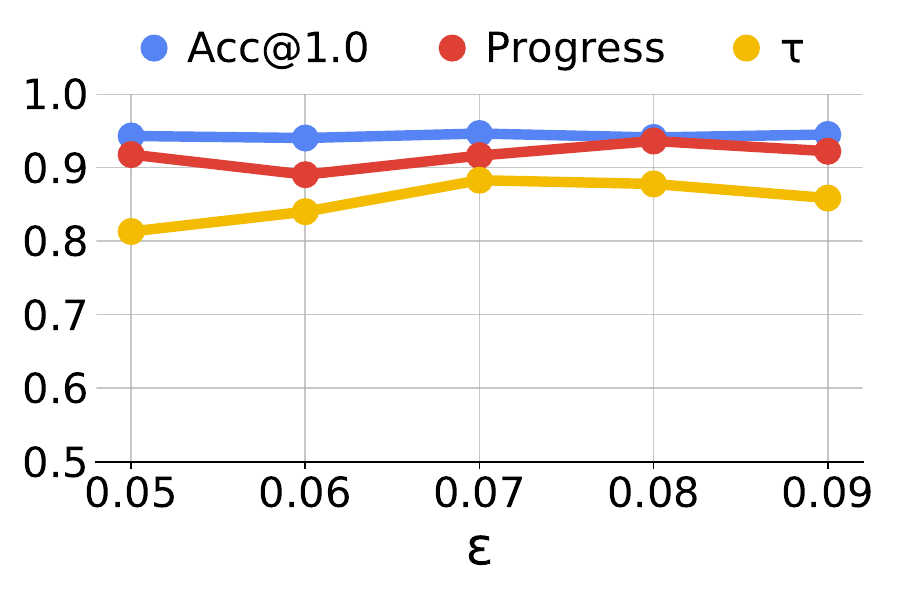}
     \caption{Sensitivity analysis results.}
     \label{fig:sens_eps}
\end{figure}

Sec.~4.2 of the main paper performs sensitivity analyses on  different hyperparameters such as $r$, $\alpha$, $\rho$, $\zeta$, $w_{seg}$, and $w_{align}$. We now plot the results of the entropy regularization weight $\epsilon$ in Fig.~\ref{fig:sens_eps}. We use our VAOT approach and the Pouring dataset for this experiment. From the results, Acc@1.0 remains stable across the studied value range of $\epsilon$. Progress exhibits small fluctuations, whereas $\tau$ is the most sensitive metric, steadily increasing from $\epsilon = 0.05$, peaking at $\epsilon = 0.07$, and declining thereafter. Furthermore, we find that $\epsilon = 0.07$ also yields the best results for the remaining datasets.

\subsection{Multi-Action Video Alignment Results}
\label{sec:supp_multi_action}

\input{Tables/supp_multiaction_alignment_results}

In Sec.~4.3 of the main paper, we train a separate encoder for each action of the Penn Action dataset and report the average results across all $13$ actions of the Penn Action dataset, which is not efficient both in terms of time and memory consumption. In this section, we train only a single encoder for all actions of Penn Action and report the multi-action video alignment results in Tab.~\ref{tab:multiaction_alignment_results}. This is a challenging experiment setting since the shared encoder needs to jointly extract useful features for all actions. We use our VAOT approach for this experiment. It is evident from Tab.~\ref{tab:multiaction_alignment_results} that our VAOT approach achieves the best results across all metrics, outperforming all competing methods in this experiment setting. Especially, on Progress, VAOT outperforms previous works by significant margins.

\subsection{Per-Video Action Segmentation Results}
\label{sec:supp_per_video}

\input{Tables/supp_per_video_seg}

We perform Hungarian matching over the entire dataset to obtain the full-dataset action segmentation results in Tab.~3 of the main paper. In the following, we benchmark our VASOT approach against previous unsupervised action segmentation methods~\cite{sarfraz2021temporally,du2022fast} which conduct Hungarian matching and evaluate per video. Per-video matching and evaluation do not require clusters across all videos and hence tend to yield better results. Tab.~\ref{tab:segmentation_results_per_video} presents the results in the per-video matching and evaluation setting. It can be seen from the results that our VASOT approach achieves the best overall performance in this experiment setting. Especially, on F1 score, our VASOT approach consistently performs the best across all datasets.

\subsection{Qualitative Results}
\label{sec:supp_qualitative}

\begin{figure}[t]
    \centering
    \includegraphics[width=0.48\textwidth, trim = 5mm 24mm 5mm 26mm, clip]{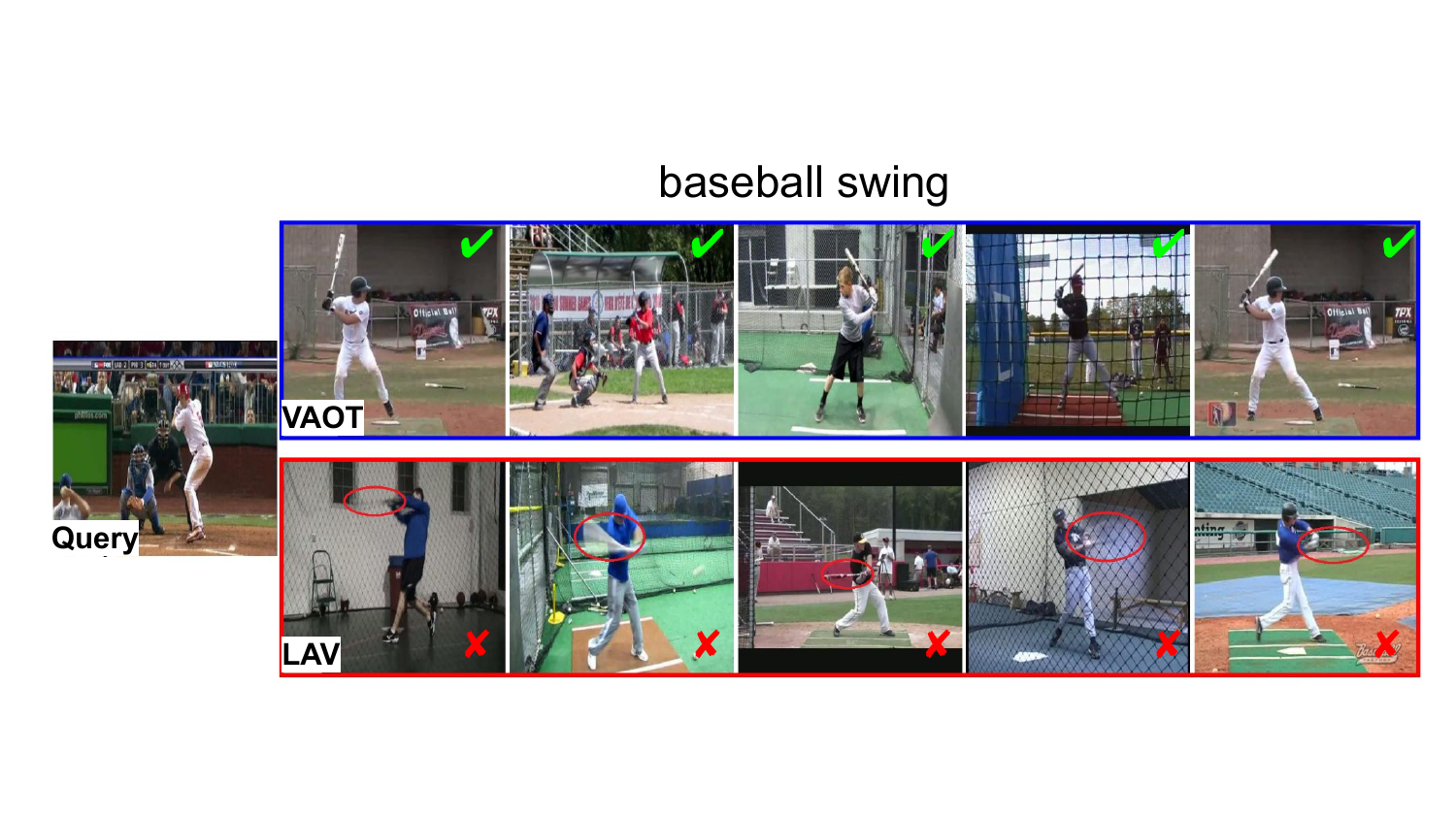} 
    \caption{Fine-grained frame retrieval results on Penn Action. The query image is on the left, while on the right are the top 5 matching images retrieved by VAOT (blue box) and LAV (red box).}
    \label{fig:supp_frame_retrieval}
\end{figure}

\begin{figure}[t]
     \centering
     \begin{subfigure}[t]{0.23\textwidth}
         \centering
         \includegraphics[width=\textwidth]{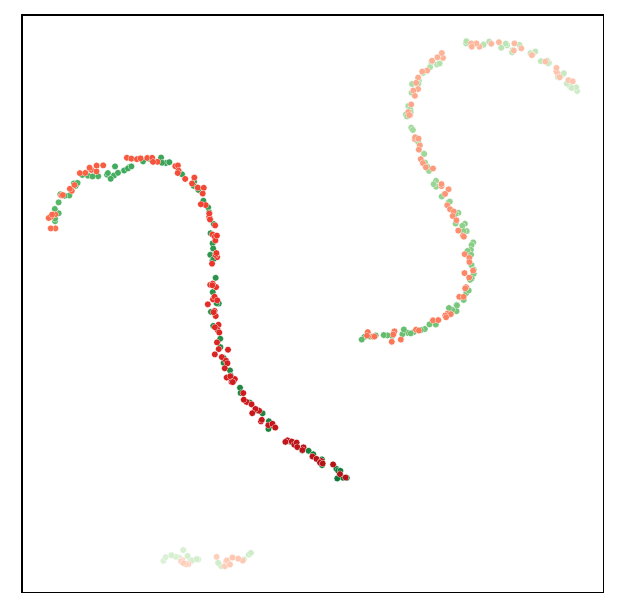}
         \caption{LAV}
         \label{fig:alignment_tsne(LAV)}
     \end{subfigure}
      \begin{subfigure}[t]{0.23\textwidth}
         \centering
         \includegraphics[width=\textwidth]{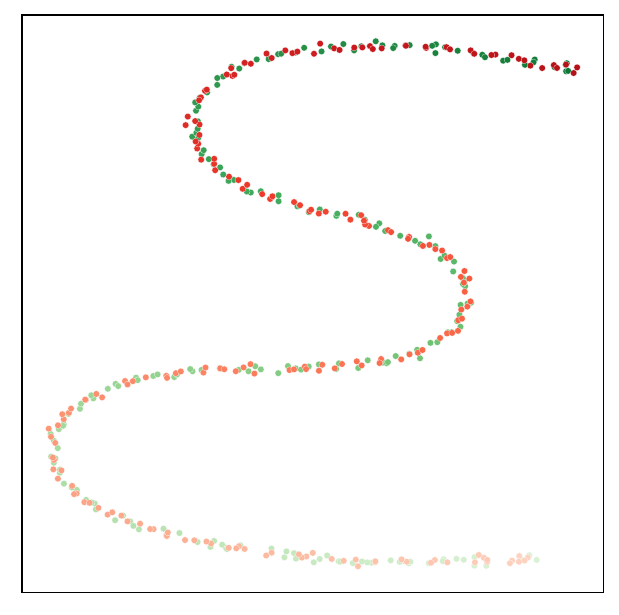}
         \caption{VAOT (Ours)}
         \label{fig:alignment_tsne(VAOT)}
     \end{subfigure}
    \caption{t-SNE visualizations of learned frame embeddings of two Pouring videos (green and red). The color opacity represents the temporal frame index from the first frame to the last frame.}
    \label{fig:Alignment_tsne}
\end{figure}

\begin{figure}[t]
    \centering
    \includegraphics[width=0.48\textwidth, trim = 0mm 0mm 0mm 0mm, clip]{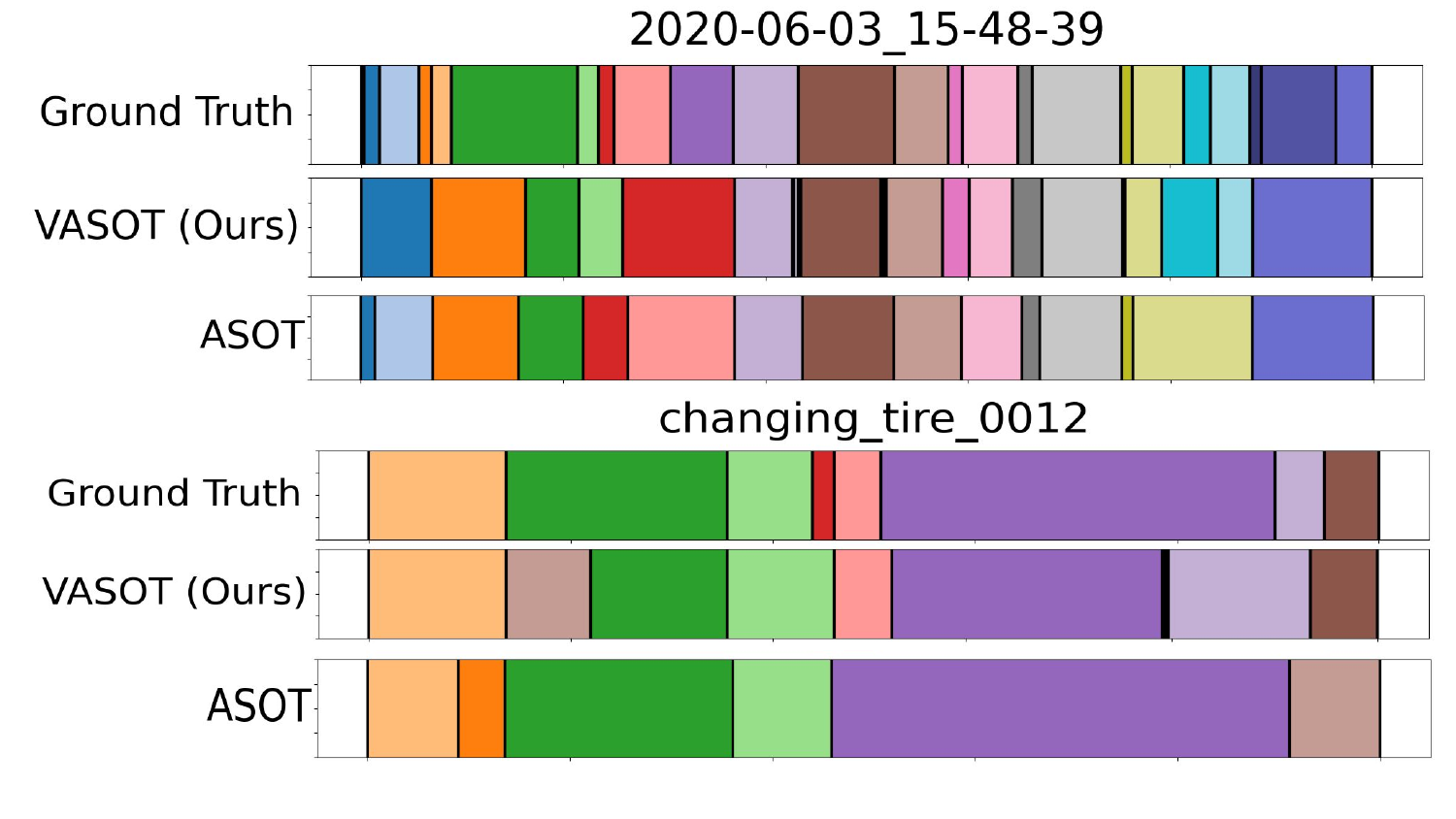} 
    \caption{Action segmentation results on Desktop Assembly (top) and YouTube Instructions (bottom).}
    \label{fig:segmentation2}
\end{figure}

We provide some qualitative results in this section. Firstly, Fig.~\ref{fig:supp_frame_retrieval} presents the frame retrieval results of our VAOT approach and LAV~\cite{haresh2021learning} on Penn Action. From Fig.~\ref{fig:supp_frame_retrieval}, our VAOT approach retrieves all 5 correct frames with the same action (\emph{Bat swung back fully}) as the query image, whereas LAV~\cite{haresh2021learning} retrieves all incorrect frames (\emph{Bat hits ball}), highlighted by red ovals.

Secondly, Fig.~\ref{fig:Alignment_tsne} illustrates the t-SNE visualizations of the learned frame embeddings of two Pouring videos by our VAOT approach and LAV~\cite{haresh2021learning}. In Fig.~\ref{fig:Alignment_tsne}(a), the embeddings by LAV~\cite{haresh2021learning} form locally continuous but globally fragmented trajectories, with visible gaps in earlier and later frames. In contrast, for our VAOT approach in Fig.~\ref{fig:Alignment_tsne}(b), the embeddings of corresponding frames from both videos are spatially closer and follow smoother trajectories. This suggests that our VAOT approach can effectively capture both local and global temporal information, resulting in more temporally consistent embeddings.

Lastly, Fig.~\ref{fig:segmentation2} shows the action segmentation results by our VASOT approach and ASOT~\cite{xu2024temporally} on a Desktop Assembly video and a YouTube Instructions video. The Desktop Assembly dataset is relatively balanced, with actions of more uniform durations, whereas the YouTube Instructions dataset is more unbalanced, with some actions being significantly longer or shorter than others. From Fig.~\ref{fig:segmentation2}, our VASOT approach can effectively handle both cases, yielding segmentations which align more closely with ground truth than ASOT~\cite{xu2024temporally}.

\subsection{Complexity Comparison Results}
\label{sec:supp_complexity}

We compare the complexity (in terms of model size and training time) of our multi-task VASOT approach and the separate single-task models (VAOT+ASOT) on an Nvidia 3090Ti GPU. Using a ResNet-50 encoder on Pouring, VASOT needs ($108$ MB, $116$ mins) vs. ($216$ MB, $162$ mins) of VAOT+ASOT. Using an MLP encoder on Desktop Assembly, VASOT needs ($287$ KB, $15$ mins) vs. ($571$ KB, $21$ mins) of VAOT+ASOT. The results validate that our multi-task VASOT approach saves both memory and training time as compared to the separate single-task models (VAOT+ASOT).

%% file: Tables/supp_alignment_hyperparams.tex
\begin{table*}[t]
\begin{minipage}{\linewidth}
\centering
\footnotesize
{
\begin{tabular}{c|c|c}
\specialrule{1pt}{1pt}{1pt}
\textbf{Hyperparameter} & \textbf{VAOT} & \textbf{VASOT} \\
\midrule
Number of sampled frames & 40 (P), 20 (PA, IA) & 40 (P), 20 (PA, IA) \\
Learning rate & $10^{-5}$ & $10^{-4}$ (P), $10^{-5}$ (PA, IA) \\
Weight decay & $10^{-5}$ & $10^{-5}$ \\
Batch size & 2 videos & 2 videos \\
Entropy regularization weight $\epsilon$ & 0.07 & 0.07 \\
Virtual frame threshold $\zeta$ & 0.5 & 0.5 \\
Gromov-Wasserstein weight $\alpha$ & 0.3 & 0.3 \\
Structural prior radius $r$ & 0.02 & 0.02 \\
Temporal prior weight $\rho$ & 0.35 & 0.35 \\
\specialrule{1pt}{1pt}{1pt}
\end{tabular}
}
\caption{Hyperparameter settings of our VAOT and VASOT approaches for state-of-the-art video alignment comparison. Note that P, PA, and IA represent Pouring, Penn Action, and IKEA ASM respectively.}
\label{tab:hyperparams}
\end{minipage}
\end{table*}

%% file: Tables/supp_seg_hyperparameters.tex
\begin{table*}[t]
\begin{minipage}{\linewidth}
\centering
\footnotesize
{
\begin{tabular}{c|c}
\specialrule{1pt}{1pt}{1pt}
\textbf{Hyperparameter} &  \textbf{VASOT} \\
\midrule
Number of sampled frames & 256 \\
Learning rate & $10^{-3}$ (B, M, E, DA), $10^{-4}$ (YTI) \\
Weight decay & $10^{-4}$ (B, M, E, DA), $10^{-5}$ (YTI) \\
Batch size & 2 videos \\
Entropy regularization weight $\epsilon$ & 0.07 \\
Virtual frame threshold $\zeta$ & 0.5 \\
Gromov-Wasserstein weight $\alpha$ & 0.3 (YTI, M, E, DA), 0.5 (B) \\
Structural prior radius $r$ & 0.02 (DA), 0.04 (B, YTI, M, E) \\
Temporal prior weight $\rho$ & 0.15 (M, E), 0.2 (B, YTI, DA) \\
Number of epochs & 30 (YTI, E), 50 (B), 100 (M, DA)\\
\specialrule{1pt}{1pt}{1pt}
\end{tabular}
}
\caption{Hyperparameter settings of our VASOT approach for state-of-the-art action segmentation comparison. Note that B, YTI, M, E, and DA denotes Breakfast, YouTube Instructions, 50 Salads (Mid), 50 Salads (Eval), and Desktop Assembly respectively.}
\label{tab:seg_hyperparams}
\end{minipage}
\end{table*}

%% file: Tables/supp_ablation_results.tex
\begin{table*}[!ht]
\begin{minipage}{\linewidth}
\centering
\footnotesize
{
\begin{tabular}{c|c|c|c|c|c|c|c|c|c}
\specialrule{1pt}{1pt}{1pt}
& \textbf{Method} & \textbf{Acc@0.1} & \textbf{Acc@0.5} & \textbf{Acc@1.0} & \textbf{Progress} & \bm{$\tau$} & \textbf{AP@5} & \textbf{AP@10} & \textbf{AP@15} \\

\midrule
\multirow{5}{*}{\rotatebox[origin=c]{90}{\textbf{Pouring}}}
&w/o Structural Prior
& 62.13 & 88.28 & \underline{93.68} & \underline{90.28} & 72.49 & 84.45 & 84.45 & 84.41\\
&w/o Temporal Prior
& 71.57 & \underline{90.51} & 91.47 & 86.94 & 78.11 & 86.17 & 86.29 & 86.24\\
&w/o Balanced Assignment
& 63.29 & 88.57 & 92.38 & 85.73 & 68.78 & 82.83 & 82.71 & 82.49\\
&w/o Virtual Frame
& \underline{86.32} & 87.22 & 93.24 & 88.11 & \underline{82.65} & \underline{86.99} & \underline{86.74} & \underline{86.56}\\
&All
& \textbf{91.80} & \textbf{92.88} & \textbf{94.63} & \textbf{91.63} & \textbf{88.28} & \textbf{91.34} & \textbf{90.56} & \textbf{90.29}\\

\specialrule{1pt}{1pt}{1pt}
\end{tabular}
}
\caption{Ablation analysis results. \textbf{Bold} and \underline{underline} denote the best and second best respectively.}
\label{tab:pouring_ablation_results}
\end{minipage}
\end{table*}

%% file: Tables/supp_multiaction_alignment_results.tex
\begin{table*}[!ht]
\begin{minipage}{\linewidth}
\centering
\footnotesize
{
\begin{tabular}{c|c|c|c|c}
\specialrule{1pt}{1pt}{1pt}
& \textbf{Method} & \textbf{Acc@1.0} & \textbf{Progress} & \bm{$\tau$} \\

\midrule
\multirow{7}{*}{\rotatebox[origin=c]{90}{\textbf{Penn Action}}}
&SAL~\cite{misra2016shuffle}
& 68.15 & 39.03 & 47.44\\
&TCN~\cite{sermanet2018time}
& 68.09 & 38.34 & 54.17\\
&TCC~\cite{dwibedi2019tcc}
& 74.39 & 59.14 & 64.08\\
&LAV~\cite{haresh2021learning}
& 78.68 & 62.52 & 68.35\\
&VAVA~\cite{liu2022learning}
& \underline{80.25} & 64.82 & \underline{76.20}\\
&GTCC~\cite{donahue2024learning}
& 73.90 & \underline{69.70} & 60.70\\
&\cellcolor{beaublue} VAOT (Ours)
&\cellcolor{beaublue} \textbf{83.49} &\cellcolor{beaublue} \textbf{79.23} &\cellcolor{beaublue} \textbf{77.68}\\

\specialrule{1pt}{1pt}{1pt}
\end{tabular}
}
\caption{Multi-action video alignment results. \textbf{Bold} and \underline{underline} denote the best and second best respectively.}
\label{tab:multiaction_alignment_results}
\end{minipage}
\end{table*}

%% file: Tables/supp_per_video_seg.tex
\begin{table*}[!ht]
\centering
\footnotesize
\begin{tabular}{c|c|c|c|c|c|c}
\toprule
& \multirow{2}{*}{\textbf{Method}} & \textbf{Breakfast} & \textbf{YouTube Instructions} & \textbf{50 Salads (Mid)} & \textbf{50 Salads (Eval)} & \textbf{Desktop Assembly} \\
\cmidrule(lr){3-3} \cmidrule(lr){4-4} \cmidrule(lr){5-5} \cmidrule(lr){6-6} \cmidrule(lr){7-7}
& & \textbf{MoF / F1 / mIoU} & \textbf{MoF / F1 / mIoU} & \textbf{MoF / F1 / mIoU} & \textbf{MoF / F1 / mIoU} & \textbf{MoF / F1 / mIoU} \\

\midrule
\multirow{4}{*}{\rotatebox[origin=c]{90}{\textbf{Per-Video}}} & TWF~\cite{sarfraz2021temporally} & 62.7 / 49.8 / \textbf{42.3} & 56.7 / 48.2 / - & \underline{66.8} / \underline{56.4} / \textbf{48.7} & \textbf{71.7} / - / - & \textbf{73.3} / 67.7 / \textbf{57.7} \\
& ABD~\cite{du2022fast} & \underline{64.0} / 52.3 / - & 67.2 / 49.2 / - & \textbf{71.8} / - / - & \underline{71.22}/ - / - & - / - / - \\
& ASOT~\cite{xu2024temporally} & 63.3 / \underline{53.5} / \underline{35.9} & \textbf{71.2} / \underline{63.3} / \underline{47.8} & 64.3 / 51.1 / 33.4 & 64.5 / \underline{58.9} / \underline{33.0} & \underline{73.0} / 68.4 / 47.6 \\
&\cellcolor{beaublue} VASOT (Ours) &\cellcolor{beaublue} \textbf{64.5} / \textbf{54.3} / 35.8 &\cellcolor{beaublue} \underline{70.1} / \textbf{67.5} / \textbf{53.0} &\cellcolor{beaublue} 64.7 / \textbf{64.2} / \underline{45.1} &\cellcolor{beaublue} 55.1 / \textbf{59.5} / \textbf{37.7} &\cellcolor{beaublue} 72.1 / \textbf{77.2} / \underline{53.2} \\
        
\bottomrule
\end{tabular}
\caption{Per-video action segmentation results. \textbf{Bold} and \underline{underline} denote the best and second best respectively.}
\label{tab:segmentation_results_per_video}
\end{table*}

%% file: references.bib
@String(CVPR= {IEEE Conf. Comput. Vis. Pattern Recog.})

@String(ECCV= {Eur. Conf. Comput. Vis.})

@String(NIPS= {Adv. Neural Inform. Process. Syst.})

@String(ICIP = {IEEE Int. Conf. Image Process.})

@String(AAAI = {AAAI})

@String(CVPR  = {CVPR})

@String(ECCV  = {ECCV})

@String(NIPS  = {NeurIPS})

@String(ICIP  = {ICIP})

@inproceedings{larsson2017colorization,
  title={Colorization as a proxy task for visual understanding},
  author={Larsson, Gustav and Maire, Michael and Shakhnarovich, Gregory},
  booktitle={Proceedings of the IEEE Conference on Computer Vision and Pattern Recognition},
  pages={6874--6883},
  year={2017}
}

@inproceedings{larsson2016learning,
  title={Learning representations for automatic colorization},
  author={Larsson, Gustav and Maire, Michael and Shakhnarovich, Gregory},
  booktitle={European conference on computer vision},
  pages={577--593},
  year={2016},
  organization={Springer}
}

@inproceedings{noroozi2017representation,
  title={Representation learning by learning to count},
  author={Noroozi, Mehdi and Pirsiavash, Hamed and Favaro, Paolo},
  booktitle={Proceedings of the IEEE International Conference on Computer Vision},
  pages={5898--5906},
  year={2017}
}

@inproceedings{liu2018leveraging,
  title={Leveraging unlabeled data for crowd counting by learning to rank},
  author={Liu, Xialei and Van De Weijer, Joost and Bagdanov, Andrew D},
  booktitle={Proceedings of the IEEE Conference on Computer Vision and Pattern Recognition},
  pages={7661--7669},
  year={2018}
}

@inproceedings{kim2019self,
  title={Self-supervised video representation learning with space-time cubic puzzles},
  author={Kim, Dahun and Cho, Donghyeon and Kweon, In So},
  booktitle={Proceedings of the AAAI Conference on Artificial Intelligence},
  volume={33},
  pages={8545--8552},
  year={2019}
}

@inproceedings{carlucci2019domain,
  title={Domain generalization by solving jigsaw puzzles},
  author={Carlucci, Fabio M and D'Innocente, Antonio and Bucci, Silvia and Caputo, Barbara and Tommasi, Tatiana},
  booktitle={Proceedings of the IEEE Conference on Computer Vision and Pattern Recognition},
  pages={2229--2238},
  year={2019}
}

@inproceedings{
gidaris2018unsupervised,
title={Unsupervised Representation Learning by Predicting Image Rotations},
author={Spyros Gidaris and Praveer Singh and Nikos Komodakis},
booktitle={International Conference on Learning Representations},
year={2018},
url={https://openreview.net/forum?id=S1v4N2l0-},
}

@inproceedings{jenni2020steering,
  title={Steering self-supervised feature learning beyond local pixel statistics},
  author={Jenni, Simon and Jin, Hailin and Favaro, Paolo},
  booktitle={Proceedings of the IEEE/CVF Conference on Computer Vision and Pattern Recognition},
  pages={6408--6417},
  year={2020}
}

@inproceedings{caron2018deep,
  title={Deep clustering for unsupervised learning of visual features},
  author={Caron, Mathilde and Bojanowski, Piotr and Joulin, Armand and Douze, Matthijs},
  booktitle={Proceedings of the European Conference on Computer Vision (ECCV)},
  pages={132--149},
  year={2018}
}

@inproceedings{caron2019unsupervised,
  title={Unsupervised pre-training of image features on non-curated data},
  author={Caron, Mathilde and Bojanowski, Piotr and Mairal, Julien and Joulin, Armand},
  booktitle={Proceedings of the IEEE/CVF International Conference on Computer Vision},
  pages={2959--2968},
  year={2019}
}

@inproceedings{srivastava2015unsupervised,
  title={Unsupervised learning of video representations using lstms},
  author={Srivastava, Nitish and Mansimov, Elman and Salakhudinov, Ruslan},
  booktitle={International conference on machine learning},
  pages={843--852},
  year={2015}
}

@inproceedings{vondrick2016generating,
  title={Generating videos with scene dynamics},
  author={Vondrick, Carl and Pirsiavash, Hamed and Torralba, Antonio},
  booktitle={Advances in neural information processing systems},
  pages={613--621},
  year={2016}
}

@article{ahsan2018discrimnet,
  title={Discrimnet: Semi-supervised action recognition from videos using generative adversarial networks},
  author={Ahsan, Unaiza and Sun, Chen and Essa, Irfan},
  journal={arXiv preprint},
  year={2018}
}

@inproceedings{diba2019dynamonet,
  title={Dynamonet: Dynamic action and motion network},
  author={Diba, Ali and Sharma, Vivek and Gool, Luc Van and Stiefelhagen, Rainer},
  booktitle={Proceedings of the IEEE International Conference on Computer Vision},
  pages={6192--6201},
  year={2019}
}

@inproceedings{mobahi2009deep,
  title={Deep learning from temporal coherence in video},
  author={Mobahi, Hossein and Collobert, Ronan and Weston, Jason},
  booktitle={Proceedings of the 26th Annual International Conference on Machine Learning},
  pages={737--744},
  year={2009}
}

@inproceedings{zou2011unsupervised,
  title={Unsupervised learning of visual invariance with temporal coherence},
  author={Zou, Will Y and Ng, Andrew Y and Yu, Kai},
  booktitle={NIPS 2011 workshop on deep learning and unsupervised feature learning},
  volume={3},
  year={2011}
}

@inproceedings{goroshin2015unsupervised,
  title={Unsupervised learning of spatiotemporally coherent metrics},
  author={Goroshin, Ross and Bruna, Joan and Tompson, Jonathan and Eigen, David and LeCun, Yann},
  booktitle={Proceedings of the IEEE international conference on computer vision},
  pages={4086--4093},
  year={2015}
}

@inproceedings{misra2016shuffle,
  title={Shuffle and learn: unsupervised learning using temporal order verification},
  author={Misra, Ishan and Zitnick, C Lawrence and Hebert, Martial},
  booktitle={European Conference on Computer Vision},
  pages={527--544},
  year={2016},
  organization={Springer}
}

@inproceedings{lee2017unsupervised,
  title={Unsupervised representation learning by sorting sequences},
  author={Lee, Hsin-Ying and Huang, Jia-Bin and Singh, Maneesh and Yang, Ming-Hsuan},
  booktitle={Proceedings of the IEEE International Conference on Computer Vision},
  pages={667--676},
  year={2017}
}

@inproceedings{fernando2017self,
  title={Self-supervised video representation learning with odd-one-out networks},
  author={Fernando, Basura and Bilen, Hakan and Gavves, Efstratios and Gould, Stephen},
  booktitle={Proceedings of the IEEE conference on computer vision and pattern recognition},
  pages={3636--3645},
  year={2017}
}

@inproceedings{xu2019self,
  title={Self-supervised spatiotemporal learning via video clip order prediction},
  author={Xu, Dejing and Xiao, Jun and Zhao, Zhou and Shao, Jian and Xie, Di and Zhuang, Yueting},
  booktitle={Proceedings of the IEEE Conference on Computer Vision and Pattern Recognition},
  pages={10334--10343},
  year={2019}
}

@inproceedings{pickup2014seeing,
  title={Seeing the arrow of time},
  author={Pickup, Lyndsey C and Pan, Zheng and Wei, Donglai and Shih, YiChang and Zhang, Changshui and Zisserman, Andrew and Scholkopf, Bernhard and Freeman, William T},
  booktitle={Proceedings of the IEEE Conference on Computer Vision and Pattern Recognition},
  pages={2035--2042},
  year={2014}
}

@inproceedings{wei2018learning,
  title={Learning and using the arrow of time},
  author={Wei, Donglai and Lim, Joseph J and Zisserman, Andrew and Freeman, William T},
  booktitle={Proceedings of the IEEE conference on computer vision and pattern recognition},
  pages={8052--8060},
  year={2018}
}

@inproceedings{benaim2020speednet,
  title={Speednet: Learning the speediness in videos},
  author={Benaim, Sagie and Ephrat, Ariel and Lang, Oran and Mosseri, Inbar and Freeman, William T and Rubinstein, Michael and Irani, Michal and Dekel, Tali},
  booktitle={Proceedings of the IEEE/CVF Conference on Computer Vision and Pattern Recognition},
  pages={9922--9931},
  year={2020}
}

@inproceedings{wang2020self,
  title={Self-supervised video representation learning by pace prediction},
  author={Wang, Jiangliu and Jiao, Jianbo and Liu, Yun-Hui},
  booktitle={Computer Vision--ECCV 2020: 16th European Conference, Glasgow, UK, August 23--28, 2020, Proceedings, Part XVII 16},
  pages={504--521},
  year={2020},
  organization={Springer}
}

@inproceedings{yao2020video,
  title={Video playback rate perception for self-supervised spatio-temporal representation learning},
  author={Yao, Yuan and Liu, Chang and Luo, Dezhao and Zhou, Yu and Ye, Qixiang},
  booktitle={Proceedings of the IEEE/CVF conference on computer vision and pattern recognition},
  pages={6548--6557},
  year={2020}
}

@inproceedings{feichtenhofer2021large,
  title={A large-scale study on unsupervised spatiotemporal representation learning},
  author={Feichtenhofer, Christoph and Fan, Haoqi and Xiong, Bo and Girshick, Ross and He, Kaiming},
  booktitle={Proceedings of the IEEE/CVF Conference on Computer Vision and Pattern Recognition},
  pages={3299--3309},
  year={2021}
}

@inproceedings{hu2021contrast,
  title={Contrast and order representations for video self-supervised learning},
  author={Hu, Kai and Shao, Jie and Liu, Yuan and Raj, Bhiksha and Savvides, Marios and Shen, Zhiqiang},
  booktitle={Proceedings of the IEEE/CVF International Conference on Computer Vision},
  pages={7939--7949},
  year={2021}
}

@inproceedings{qian2021spatiotemporal,
  title={Spatiotemporal contrastive video representation learning},
  author={Qian, Rui and Meng, Tianjian and Gong, Boqing and Yang, Ming-Hsuan and Wang, Huisheng and Belongie, Serge and Cui, Yin},
  booktitle={Proceedings of the IEEE/CVF Conference on Computer Vision and Pattern Recognition},
  pages={6964--6974},
  year={2021}
}

@article{dave2022tclr,
  title={Tclr: Temporal contrastive learning for video representation},
  author={Dave, Ishan and Gupta, Rohit and Rizve, Mamshad Nayeem and Shah, Mubarak},
  journal={Computer Vision and Image Understanding},
  volume={219},
  pages={103406},
  year={2022},
  publisher={Elsevier}
}

@inproceedings{zheng2018unsupervised,
  title={Unsupervised representation learning with long-term dynamics for skeleton based action recognition},
  author={Zheng, Nenggan and Wen, Jun and Liu, Risheng and Long, Liangqu and Dai, Jianhua and Gong, Zhefeng},
  booktitle={Proceedings of the AAAI Conference on Artificial Intelligence},
  volume={32},
  number={1},
  year={2018}
}

@inproceedings{su2020predict,
  title={Predict \& cluster: Unsupervised skeleton based action recognition},
  author={Su, Kun and Liu, Xiulong and Shlizerman, Eli},
  booktitle={Proceedings of the IEEE/CVF Conference on Computer Vision and Pattern Recognition},
  pages={9631--9640},
  year={2020}
}

@inproceedings{kwon2022context,
  title={Context-aware sequence alignment using 4d skeletal augmentation},
  author={Kwon, Taein and Tekin, Bugra and Tang, Siyu and Pollefeys, Marc},
  booktitle={Proceedings of the IEEE/CVF Conference on Computer Vision and Pattern Recognition},
  pages={8172--8182},
  year={2022}
}

@inproceedings{tran2024learning,
  title={Learning by Aligning 2D Skeleton Sequences and Multi-modality Fusion},
  author={Tran, Quoc-Huy and Ahmed, Muhammad and Popattia, Murad and Ahmed, M Hassan and Konin, Andrey and Zia, M Zeeshan},
  booktitle={European Conference on Computer Vision},
  pages={141--161},
  year={2024},
  organization={Springer}
}

@inproceedings{si2020adversarial,
  title={Adversarial self-supervised learning for semi-supervised 3d action recognition},
  author={Si, Chenyang and Nie, Xuecheng and Wang, Wei and Wang, Liang and Tan, Tieniu and Feng, Jiashi},
  booktitle={Computer Vision--ECCV 2020: 16th European Conference, Glasgow, UK, August 23--28, 2020, Proceedings, Part VII 16},
  pages={35--51},
  year={2020},
  organization={Springer}
}

@inproceedings{su2021self,
  title={Self-supervised 3d skeleton action representation learning with motion consistency and continuity},
  author={Su, Yukun and Lin, Guosheng and Wu, Qingyao},
  booktitle={Proceedings of the IEEE/CVF international conference on computer vision},
  pages={13328--13338},
  year={2021}
}

@inproceedings{lin2020ms2l,
  title={Ms2l: Multi-task self-supervised learning for skeleton based action recognition},
  author={Lin, Lilang and Song, Sijie and Yang, Wenhan and Liu, Jiaying},
  booktitle={Proceedings of the 28th ACM International Conference on Multimedia},
  pages={2490--2498},
  year={2020}
}

@InProceedings{dwibedi2019tcc,
author = {Dwibedi, Debidatta and Aytar, Yusuf and Tompson, Jonathan and Sermanet, Pierre and Zisserman, Andrew},
title = {Temporal Cycle-Consistency Learning},
booktitle = {Proceedings of the IEEE/CVF Conference on Computer Vision and Pattern Recognition (CVPR)},
month = {June},
year = {2019}
}

@inproceedings{donahue2024learning,
  title={Learning to predict activity progress by self-supervised video alignment},
  author={Donahue, Gerard and Elhamifar, Ehsan},
  booktitle={Proceedings of the IEEE/CVF Conference on Computer Vision and Pattern Recognition},
  pages={18667--18677},
  year={2024}
}

@inproceedings{haresh2021learning,
  title={Learning by aligning videos in time},
  author={Haresh, Sanjay and Kumar, Sateesh and Coskun, Huseyin and Syed, Shahram N and Konin, Andrey and Zia, Zeeshan and Tran, Quoc-Huy},
  booktitle={Proceedings of the IEEE/CVF Conference on Computer Vision and Pattern Recognition},
  pages={5548--5558},
  year={2021}
}

@inproceedings{liu2022learning,
  title={Learning to align sequential actions in the wild},
  author={Liu, Weizhe and Tekin, Bugra and Coskun, Huseyin and Vineet, Vibhav and Fua, Pascal and Pollefeys, Marc},
  booktitle={Proceedings of the IEEE/CVF Conference on Computer Vision and Pattern Recognition},
  pages={2181--2191},
  year={2022}
}

@inproceedings{cuturi2017soft,
  title={Soft-dtw: a differentiable loss function for time-series},
  author={Cuturi, Marco and Blondel, Mathieu},
  booktitle={International conference on machine learning},
  pages={894--903},
  year={2017},
  organization={PMLR}
}

@article{cuturi2013sinkhorn,
  title={Sinkhorn distances: Lightspeed computation of optimal transport},
  author={Cuturi, Marco},
  journal={Advances in neural information processing systems},
  volume={26},
  year={2013}
}

@inproceedings{kukleva2019unsupervised,
  title={Unsupervised learning of action classes with continuous temporal embedding},
  author={Kukleva, Anna and Kuehne, Hilde and Sener, Fadime and Gall, Jurgen},
  booktitle={Proceedings of the IEEE/CVF Conference on Computer Vision and Pattern Recognition},
  pages={12066--12074},
  year={2019}
}

@inproceedings{vidalmata2021joint,
  title={Joint visual-temporal embedding for unsupervised learning of actions in untrimmed sequences},
  author={VidalMata, Rosaura G and Scheirer, Walter J and Kukleva, Anna and Cox, David and Kuehne, Hilde},
  booktitle={Proceedings of the IEEE/CVF Winter Conference on Applications of Computer Vision},
  pages={1238--1247},
  year={2021}
}

@inproceedings{li2021action,
  title={Action shuffle alternating learning for unsupervised action segmentation},
  author={Li, Jun and Todorovic, Sinisa},
  booktitle={Proceedings of the IEEE/CVF Conference on Computer Vision and Pattern Recognition},
  pages={12628--12636},
  year={2021}
}

@inproceedings{swetha2021unsupervised,
  title={Unsupervised discriminative embedding for sub-action learning in complex activities},
  author={Swetha, Sirnam and Kuehne, Hilde and Rawat, Yogesh S and Shah, Mubarak},
  booktitle={2021 IEEE International Conference on Image Processing (ICIP)},
  pages={2588--2592},
  year={2021},
  organization={IEEE}
}

@inproceedings{kumar2022unsupervised,
  title={Unsupervised action segmentation by joint representation learning and online clustering},
  author={Kumar, Sateesh and Haresh, Sanjay and Ahmed, Awais and Konin, Andrey and Zia, M Zeeshan and Tran, Quoc-Huy},
  booktitle={Proceedings of the IEEE/CVF Conference on Computer Vision and Pattern Recognition},
  pages={20174--20185},
  year={2022}
}

@inproceedings{tran2024permutation,
  title={Permutation-aware activity segmentation via unsupervised frame-to-segment alignment},
  author={Tran, Quoc-Huy and Mehmood, Ahmed and Ahmed, Muhammad and Naufil, Muhammad and Zafar, Anas and Konin, Andrey and Zia, Zeeshan},
  booktitle={Proceedings of the IEEE/CVF Winter Conference on Applications of Computer Vision},
  pages={6426--6436},
  year={2024}
}

@inproceedings{xu2024temporally,
  title={Temporally Consistent Unbalanced Optimal Transport for Unsupervised Action Segmentation},
  author={Xu, Ming and Gould, Stephen},
  booktitle={Proceedings of the IEEE/CVF Conference on Computer Vision and Pattern Recognition},
  pages={14618--14627},
  year={2024}
}

@article{ding2023temporal,
  title={Temporal action segmentation: An analysis of modern techniques},
  author={Ding, Guodong and Sener, Fadime and Yao, Angela},
  journal={IEEE Transactions on Pattern Analysis and Machine Intelligence},
  year={2023},
  publisher={IEEE}
}

@article{khamis2024scalable,
  title={Scalable Optimal Transport Methods in Machine Learning: A Contemporary Survey},
  author={Khamis, Abdelwahed and Tsuchida, Russell and Tarek, Mohamed and Rolland, Vivien and Petersson, Lars},
  journal={IEEE Transactions on Pattern Analysis and Machine Intelligence},
  year={2024},
  publisher={IEEE}
}

@inproceedings{liu2020semantic,
  title={Semantic correspondence as an optimal transport problem},
  author={Liu, Yanbin and Zhu, Linchao and Yamada, Makoto and Yang, Yi},
  booktitle={Proceedings of the IEEE/CVF Conference on Computer Vision and Pattern Recognition},
  pages={4463--4472},
  year={2020}
}

@inproceedings{sarlin2020superglue,
  title={Superglue: Learning feature matching with graph neural networks},
  author={Sarlin, Paul-Edouard and DeTone, Daniel and Malisiewicz, Tomasz and Rabinovich, Andrew},
  booktitle={Proceedings of the IEEE/CVF conference on computer vision and pattern recognition},
  pages={4938--4947},
  year={2020}
}

@article{shen2021accurate,
  title={Accurate point cloud registration with robust optimal transport},
  author={Shen, Zhengyang and Feydy, Jean and Liu, Peirong and Curiale, Ariel H and San Jose Estepar, Ruben and San Jose Estepar, Raul and Niethammer, Marc},
  journal={Advances in Neural Information Processing Systems},
  volume={34},
  pages={5373--5389},
  year={2021}
}

@inproceedings{de2023unbalanced,
  title={Unbalanced optimal transport: A unified framework for object detection},
  author={De Plaen, Henri and De Plaen, Pierre-Fran{\c{c}}ois and Suykens, Johan AK and Proesmans, Marc and Tuytelaars, Tinne and Van Gool, Luc},
  booktitle={Proceedings of the IEEE/CVF Conference on Computer Vision and Pattern Recognition},
  pages={3198--3207},
  year={2023}
}

@article{lee2024sota,
  title={SOTA: Sequential Optimal Transport Approximation For Visual Tracking in Wild Scenario},
  author={Lee, Seonghak and Park, Jisoo and Timofte, Radu and Kwon, Junseok},
  journal={IEEE Access},
  year={2024},
  publisher={IEEE}
}

@inproceedings{xu2019gromov,
  title={Gromov-wasserstein learning for graph matching and node embedding},
  author={Xu, Hongteng and Luo, Dixin and Zha, Hongyuan and Duke, Lawrence Carin},
  booktitle={International conference on machine learning},
  pages={6932--6941},
  year={2019},
  organization={PMLR}
}

@article{thual2022aligning,
  title={Aligning individual brains with fused unbalanced Gromov Wasserstein},
  author={Thual, Alexis and Tran, Quang Huy and Zemskova, Tatiana and Courty, Nicolas and Flamary, R{\'e}mi and Dehaene, Stanislas and Thirion, Bertrand},
  journal={Advances in neural information processing systems},
  volume={35},
  pages={21792--21804},
  year={2022}
}

@inproceedings{carreira2017quo,
  title={Quo vadis, action recognition? a new model and the kinetics dataset},
  author={Carreira, Joao and Zisserman, Andrew},
  booktitle={proceedings of the IEEE Conference on Computer Vision and Pattern Recognition},
  pages={6299--6308},
  year={2017}
}

@inproceedings{tran2018closer,
  title={A closer look at spatiotemporal convolutions for action recognition},
  author={Tran, Du and Wang, Heng and Torresani, Lorenzo and Ray, Jamie and LeCun, Yann and Paluri, Manohar},
  booktitle={Proceedings of the IEEE conference on Computer Vision and Pattern Recognition},
  pages={6450--6459},
  year={2018}
}

@inproceedings{wang2018non,
  title={Non-local neural networks},
  author={Wang, Xiaolong and Girshick, Ross and Gupta, Abhinav and He, Kaiming},
  booktitle={Proceedings of the IEEE conference on computer vision and pattern recognition},
  pages={7794--7803},
  year={2018}
}

@inproceedings{feichtenhofer2019slowfast,
  title={Slowfast networks for video recognition},
  author={Feichtenhofer, Christoph and Fan, Haoqi and Malik, Jitendra and He, Kaiming},
  booktitle={Proceedings of the IEEE/CVF international conference on computer vision},
  pages={6202--6211},
  year={2019}
}

@inproceedings{chowdhuryopel,
  title={OPEL: Optimal Transport Guided ProcedurE Learning},
  author={Chowdhury, Sayeed Shafayet and Chandra, Soumyadeep and Roy, Kaushik},
  booktitle={The Thirty-eighth Annual Conference on Neural Information Processing Systems},
  year={2024}
}

@article{spurio2024hierarchical,
  title={Hierarchical Vector Quantization for Unsupervised Action Segmentation},
  author={Spurio, Federico and Bahrami, Emad and Francesca, Gianpiero and Gall, Juergen},
  journal={arXiv preprint arXiv:2412.17640},
  year={2024}
}

@article{thorpe2018introduction,
  title={Introduction to optimal transport},
  author={Thorpe, Matthew},
  journal={Notes of Course at University of Cambridge},
  year={2018}
}

@inproceedings{peyre2016gromov,
  title={Gromov-wasserstein averaging of kernel and distance matrices},
  author={Peyr{\'e}, Gabriel and Cuturi, Marco and Solomon, Justin},
  booktitle={International conference on machine learning},
  pages={2664--2672},
  year={2016},
  organization={PMLR}
}

@article{vayer2020fused,
  title={Fused Gromov-Wasserstein distance for structured objects},
  author={Vayer, Titouan and Chapel, Laetitia and Flamary, R{\'e}mi and Tavenard, Romain and Courty, Nicolas},
  journal={Algorithms},
  volume={13},
  number={9},
  pages={212},
  year={2020},
  publisher={MDPI}
}

@inproceedings{titouan2019optimal,
  title={Optimal transport for structured data with application on graphs},
  author={Titouan, Vayer and Courty, Nicolas and Tavenard, Romain and Flamary, R{\'e}mi},
  booktitle={International Conference on Machine Learning},
  pages={6275--6284},
  year={2019},
  organization={PMLR}
}

@inproceedings{li2017deep,
  title={Deep supervision with shape concepts for occlusion-aware 3d object parsing},
  author={Li, Chi and Zeeshan Zia, M and Tran, Quoc-Huy and Yu, Xiang and Hager, Gregory D and Chandraker, Manmohan},
  booktitle={Proceedings of the IEEE conference on computer vision and pattern recognition},
  pages={5465--5474},
  year={2017}
}

@article{li2018deep,
  title={Deep supervision with intermediate concepts},
  author={Li, Chi and Zia, M Zeeshan and Tran, Quoc-Huy and Yu, Xiang and Hager, Gregory D and Chandraker, Manmohan},
  journal={IEEE transactions on pattern analysis and machine intelligence},
  volume={41},
  number={8},
  pages={1828--1843},
  year={2018},
  publisher={IEEE}
}

@inproceedings{sermanet2018time,
  title={Time-contrastive networks: Self-supervised learning from video},
  author={Sermanet, Pierre and Lynch, Corey and Chebotar, Yevgen and Hsu, Jasmine and Jang, Eric and Schaal, Stefan and Levine, Sergey and Brain, Google},
  booktitle={2018 IEEE international conference on robotics and automation (ICRA)},
  pages={1134--1141},
  year={2018},
  organization={IEEE}
}

@inproceedings{ben2021ikea,
  title={The ikea asm dataset: Understanding people assembling furniture through actions, objects and pose},
  author={Ben-Shabat, Yizhak and Yu, Xin and Saleh, Fatemeh and Campbell, Dylan and Rodriguez-Opazo, Cristian and Li, Hongdong and Gould, Stephen},
  booktitle={Proceedings of the IEEE/CVF Winter Conference on Applications of Computer Vision},
  pages={847--859},
  year={2021}
}

@inproceedings{kuehne2014language,
  title={The language of actions: Recovering the syntax and semantics of goal-directed human activities},
  author={Kuehne, Hilde and Arslan, Ali and Serre, Thomas},
  booktitle={Proceedings of the IEEE conference on computer vision and pattern recognition},
  pages={780--787},
  year={2014}
}

@inproceedings{stein2013combining,
  title={Combining embedded accelerometers with computer vision for recognizing food preparation activities},
  author={Stein, Sebastian and McKenna, Stephen J},
  booktitle={Proceedings of the 2013 ACM international joint conference on Pervasive and ubiquitous computing},
  pages={729--738},
  year={2013}
}

@inproceedings{alayrac2016unsupervised,
  title={Unsupervised learning from narrated instruction videos},
  author={Alayrac, Jean-Baptiste and Bojanowski, Piotr and Agrawal, Nishant and Sivic, Josef and Laptev, Ivan and Lacoste-Julien, Simon},
  booktitle={Proceedings of the IEEE conference on computer vision and pattern recognition},
  pages={4575--4583},
  year={2016}
}

@inproceedings{zhang2013actemes,
  title={From actemes to action: A strongly-supervised representation for detailed action understanding},
  author={Zhang, Weiyu and Zhu, Menglong and Derpanis, Konstantinos G},
  booktitle={Proceedings of the IEEE international conference on computer vision},
  pages={2248--2255},
  year={2013}
}

@article{kingma2014adam,
  title={Adam: A method for stochastic optimization},
  author={Kingma, Diederik P and Ba, Jimmy},
  journal={arXiv preprint arXiv:1412.6980},
  year={2014}
}

@article{paszke2017automatic,
  title={Automatic differentiation in pytorch},
  author={Paszke, Adam and Gross, Sam and Chintala, Soumith and Chanan, Gregory and Yang, Edward and DeVito, Zachary and Lin, Zeming and Desmaison, Alban and Antiga, Luca and Lerer, Adam},
  year={2017}
}

@inproceedings{estrada2009appearance,
  title={Appearance-based keypoint clustering},
  author={Estrada, Francisco J and Fua, Pascal and Lepetit, Vincent and Susstrunk, Sabine},
  booktitle={2009 IEEE Conference on Computer Vision and Pattern Recognition},
  pages={1279--1286},
  year={2009},
  organization={IEEE}
}

@inproceedings{du2022fast,
  title={Fast and unsupervised action boundary detection for action segmentation},
  author={Du, Zexing and Wang, Xue and Zhou, Guoqing and Wang, Qing},
  booktitle={Proceedings of the IEEE/CVF Conference on Computer Vision and Pattern Recognition},
  pages={3323--3332},
  year={2022}
}

@inproceedings{sarfraz2021temporally,
  title={Temporally-weighted hierarchical clustering for unsupervised action segmentation},
  author={Sarfraz, Saquib and Murray, Naila and Sharma, Vivek and Diba, Ali and Van Gool, Luc and Stiefelhagen, Rainer},
  booktitle={Proceedings of the IEEE/CVF Conference on Computer Vision and Pattern Recognition},
  pages={11225--11234},
  year={2021}
}

@inproceedings{kendall2018multi,
  title={Multi-task learning using uncertainty to weigh losses for scene geometry and semantics},
  author={Kendall, Alex and Gal, Yarin and Cipolla, Roberto},
  booktitle={Proceedings of the IEEE conference on computer vision and pattern recognition},
  pages={7482--7491},
  year={2018}
}

@inproceedings{chen2018gradnorm,
  title={Gradnorm: Gradient normalization for adaptive loss balancing in deep multitask networks},
  author={Chen, Zhao and Badrinarayanan, Vijay and Lee, Chen-Yu and Rabinovich, Andrew},
  booktitle={International conference on machine learning},
  pages={794--803},
  year={2018},
  organization={PMLR}
}
